\documentclass[11pt,a4paper]{article}

\usepackage{float}
\usepackage{multirow}
\usepackage{color}
\usepackage{amsmath}
\usepackage{graphicx}
\usepackage{amssymb}
\usepackage{booktabs}
\usepackage{capt-of}
\usepackage{ulem}
\usepackage{threeparttable}
\usepackage{placeins}
\usepackage{sistyle}
\usepackage{dblfloatfix}
\usepackage{soul}
\usepackage[table]{xcolor}
\usepackage[hidelinks,breaklinks]{hyperref}
\usepackage[caption=false,font=footnotesize]{subfig}
\usepackage{fullpage}

\newcommand{\bZ}{\mathbb{Z}}
\newcommand{\bN}{\mathbb{N}}

\newcommand{\abs}[1]{\left\vert#1\right\vert}

\newcommand{\Sim}{\operatorname{S}}
\title{A Haar Wavelet-Based Perceptual Similarity Index for Image Quality Assessment}
\author{Rafael~Reisenhofer\thanks{R. Reisenhofer is with the Working Group Computational Data Analysis, Universit\"at Bremen, Fachbereich 3, Postfach 330440, 28334 Bremen, Germany (e-mail: reisenhofer@math.uni-bremen.de).}
\and Sebastian~Bosse\thanks{S. Bosse is with the Fraunhofer Heinrich Hertz Institute (Fraunhofer HHI), 10587 Berlin, Germany (e-mail: sebastian.bosse@hhi.fraunhofer.de).} \and Gitta~Kutyniok\thanks{G. Kutyniok is with the Department of Mathematics, Technische Universit\"at Berlin, 10623 Berlin, Germany (e-mail: kutyniok@math.tu-berlin.de)}\and~Thomas~Wiegand\thanks{T. Wiegand is with the Fraunhofer Heinrich Hertz Institute
(Fraunhofer HHI), 10587 Berlin, Germany, and with the Image Communication Laboratory, Berlin Institute of Technology,
10587 Berlin, Germany (e-mail: thomas.wiegand@hhi.fraunhofer.de).}}

\date{}

\begin{document}
\maketitle

\begin{abstract}
In most practical situations, the compression or transmission of images and videos creates distortions that will eventually be perceived by a human observer. Vice versa, image and video restoration techniques, such as inpainting or denoising, aim to enhance the quality of experience of human viewers. Correctly assessing the similarity between an image and an undistorted reference image as subjectively experienced by a human viewer can thus lead to significant improvements in any transmission, compression, or restoration system. This paper introduces the Haar wavelet-based perceptual similarity index (HaarPSI), a novel and computationally inexpensive similarity measure for full reference image quality assessment. The HaarPSI utilizes the coefficients obtained from a Haar wavelet decomposition to assess local similarities between two images, as well as the relative importance of image areas. The consistency of the HaarPSI with the human quality of experience was validated on four large benchmark databases containing thousands of differently distorted images. On these databases, the HaarPSI achieves higher correlations with human opinion scores than state-of-the-art full reference similarity measures like the structural similarity index (SSIM), the feature similarity index (FSIM), and the visual saliency-based index (VSI). Along with the simple computational structure and the short execution time, these experimental results suggest a high applicability of the HaarPSI in real world tasks.
\end{abstract}


\section{Introduction}
\label{sec:intro}
% images and videos are ubiquitous, images are taken everywhere
Digital images and videos are omnipresent in daily life and the importance of
visual data is still growing: According to \cite{Cisco}, by 2020, nearly a million
minutes of video content is estimated to cross the internet every second.

% qoe is important for transmission systems, iqa is a significant factor
Typically, video and image signals are intended to be ultimately viewed by
humans. For transmission or storage, most signals are compressed in
order to meet today's channel and/or storage demands. Compression as well 
as transmission errors can introduce distortions to video or image signals that are visible to human viewers. 
For evaluating or optimizing a transmission system or parts of it, e.g. by
controlling the rate-distortion trade-off of a video encoder, it is crucial to
measure the severity of distortions in a perceptually meaningful way.
% MOS values
Quality 'in a perceptually meaningful way' can only be measured reliably in
psychometric tests. In such tests, participants are asked to
rate the subjectively perceived quality of images
or videos that have previously been subject to some kind of distortion introducing processing. The quality ratings of individual participants can eventually be averaged 
to obtain a single mean opinion score (MOS) for each stimulus. 
However, although being the gold standard for assessing perceived quality such studies are expensive and time-consuming and not feasible at all for real-time 
tasks like optimizing or monitoring transmission systems. This has
been motivating research in computational image quality assessment for decades.

% different approaches for different application scenarios
Image quality assessment methods typically belong to one of three categories
with different challenges and scopes of applications: Full reference (FR)
image quality assessment approaches require and utilize the availability of a reference image. Reduced reference (RR) methods exploit a small set of features extracted from
the reference image. No reference (NR) approaches estimate the perceived quality of a possibly distorted image
solely from the image itself \cite{Lin2011}.
Unconstrained NR IQA has the notion of being the holy grail of IQA and, when successful,
essentially replicates human abilities. 
It is, however, not a feasible approach for some applications such as, for example, encoder control for video compression. An NR quality metric used for rate-distortion optimization in a video encoder would steer the optimization towards coding decisions that remove any type of noise or artifacts. However, there are videos in which noise and artifacts were intentionally added to create a certain visual effect. As an example, the reader is invited to imagine a video encoder that removes film grain from the Quentin Tarantino movie \textit{The Hateful Eight} due to the application of an NR quality metric that penalizes ''noisy'' coding decisions. Such an encoder would change a deliberate artistic decision made by the filmmakers and thus deteriorate the viewing experience.
% state of the art

% concentrate on FR
% MSE
% PSNR
The simplest FR image quality metric is the mean
squared error (MSE), which is defined as the average of the squared differences of the reference and the distorted image. Although being widely used, it
does not correlate well with perceived visual quality \cite{Gir93}.
% SSIM
More sophisticated approaches towards perceptually accurate image quality
assessments (IQA) typically follow one of three strategies. \textit{Bottom-up}
approaches explicitly model various processing mechanisms of the human visual
system (HVS), such as masking effects \cite{Watsona1997}, contrast sensitivity
\cite{daly1990}, or just-noticeable-distortion \cite{Lubin1997,
jia2006estimating} in order to assess the perceived quality of images. For
instance, the adaptivity of the HVS to the magnitude of distortions is modeled
explicitly by the concept of most apparent distortion (MAD) \cite{LaCh2010} in order to apply
two different assessment strategies for supra- and near-threshold distortions.      
 
However, the method proposed in this paper as well as most image quality
metrics developed recently follow a \textit{top-down} approach. There, general
functional properties of the HVS (considered as a black box) are assumed in
order to identify and to exploit image features corresponding to the perceived
quality. Prominent examples are the structural similarity index (SSIM)
\cite{WBSS2004}, visual information fidelity (VIF) \cite{ShBo2006}, the gradient
similarity measure (GSM) \cite{LLN2012}, spectral residual based similarity
(SR-SIM) \cite{ZhLi2012}, and the visual saliency-induced index (VSI)
\cite{ZSL2014}.   
% SSIM
The SSIM \cite{WBSS2004} aims at taking into account the sensitivity of the human
visual system towards structural information. This is done by pooling three
complementary components, namely luminance similarity (comparing local mean
luminance values), contrast similarity (comparing local variances) and
structural similarity, which is defined as the local covariance between the
reference image and its perturbed counterpart. Although being criticized
\cite{DoYa2011}, it is highly cited and among the most popular image quality
assessment metrics.  
% MSSIM
The SSIM was generalized for a multi-scale setting by the multi-scale structural
similarity index (MS-SSIM) \cite{WSB2003}. 
% VIF
One of the first information theoretic approaches to FR IQA was presented as visual information fidelity (VIF) \cite{ShBo2006}. VIF models the wavelet coefficients as Gaussian Scale Mixtures and quantifies the mutual information shared between reference and test images. The information theoretic measure of mutual information is  shown to be correlated to perceived image quality.
% GSM
% Following the basic framework of SSIM, gradients are used in the Gaussian scale mixture (GSM) model \cite{LLN2012} to capture contrast and structure. Differences in gradients are then used to estimate masking effects, while the masked gradient change is pooled with luminance differences.
Changes in contrast and structure are captured by
considering local gradients in \cite{LLN2012}, while the squared difference in
pixel values between the reference image and the distorted image is used to
measure luminance variations. This approach thus follows the basic framework of combining complementary feature maps originally introduced
in \cite{WBSS2004}. Additionally, masking effects are estimated, based
on the local gradient magnitude of the reference image and incorporated when the
two feature maps are combined.  
% SR-SIM
Spectral residual-based similarity (SR-SIM) \cite{ZhLi2012} takes into account
changes in the local horizontal and vertical gradient magnitudes. Additionally,
it incorporates changes in a spectral residual-based visual saliency estimate.  
% VSI
The visual saliency-induced index (VSI) \cite{ZSL2014} follows the same line as
SR-SIM by combining similarities in the gradient magnitude and the visual
saliency.  However, it further exploits the visual saliency map for weighting
the spatial similarity pooling. Furthermore, \cite{ZSL2014} also explores the
influence of different saliency models on the performance of the proposed image
quality measure.  
% FSIM
A combination of two feature maps is also applied successfully by the feature
similarity index (FSIM) \cite{ZZMZ2011}. Due to its conceptual similarity to
the proposed method, it will be discussed in more detail in a later section.

Adopting the advances in machine learning and data science, IQA methods
following a third, purely \textit{data driven} strategy have been proposed
recently. So far, data driven approaches were mainly developed for the domain of
NR IQA \cite{Kang2014,Ye2012,Zhang2015,bosse2016dnnNrIqa}, but they have also
been adapted in the context of FR IQA \cite{bosse2016dnnFrIqa}.   

% contributions
% What's going on here. HaarPSI is really good.

\subsection{Contributions}

This work introduces the Haar wavelet-based perceptual similarity index
(HaarPSI), a novel and computationally inexpensive measure yielding FR image
quality assessments. The HaarPSI utilizes the magnitudes of high-frequency Haar wavelet coefficients to define local similarities and low-frequency Haar wavelet coefficients to weight the importance of (dis)similarities at specific locations in the image domain.

{The six discrete two-dimensional Haar wavelet filters used in the definition of the HaarPSI respond to horizontal and vertical edges on different frequency scales. The HaarPSI is thus based on elementary implementations of functional properties known to be exhibited by neurons in the primary visual cortex, namely orientation selectivity and spatial frequency selectivity. We aim to demonstrate that such a simple model already suffices to define a similarity measure that yields state-of-the-art correlations with human opinion scores.}

The HaarPSI can also be seen as a drastic simplification of the FSIM \cite{ZZMZ2011}, which is based on a similar combination of similarity and weight maps. In the definition of the FSIM, both local similarities and weights rely on the phase congruency measure \cite{Kov2000}, whose computation requires images to be convolved with 16 complex-valued filters and contains several non-trivial steps such as adaptive thresholding. For the HaarPSI on the other hand, the two maps are computed from the responses of only six discrete Haar wavelet filters and are cleanly separated in the sense that local similarities and weights are based on different frequency scales. Surprisingly, these simplifications not only decrease the required computational effort but also lead to consistently higher correlations with human mean opinions scores.

In Section~\ref{sec:results}, we evaluate the consistency of the HaarPSI with the human
quality of experience and compare its performance to state-of-the-art similarity
measures like SSIM \cite{WBSS2004}, FSIM \cite{ZZMZ2011}, and VSI
\cite{ZSL2014}. As depicted in
Tables~\ref{tab:sroccdatabases}~and~\ref{tab:overallperf}, the HaarPSI achieves
higher correlations with human opinion scores than all other considered FR
quality metrics in all test cases except one, where it only comes second to the VSI.
In addition, the HaarPSI can be computed significantly faster than the metrics yielding the
second and third highest correlations with human opinion scores, namely VSI and
FSIM.    
In  order  to  facilitate reproducible research, our Matlab implement of the HaarPSI is publicly available
at \url{http://www.haarpsi.org/}.

% It's funny that Haar wavelets work so well
It is both convenient and surprising that the promising experimental results of the HaarPSI are based on the responses of Haar filters,
which are arguably the simplest and computationally most efficient wavelet filters existing. The results of a numerical analysis of the applicability
of other wavelet filters in the newly proposed similarity measure can be found in Table~\ref{tab:otherwavelets}.

\subsection{The Feature Similarity Index (FSIM)}
The feature similarity index (FSIM) \cite{ZZMZ2011}, proposed in 2011, is currently one of the most successful and influential FR image quality metrics. The FSIM combines two feature maps derived from the phase congruency measure \cite{Kov2000} and the local gradients of the reference and the distorted image to assess local similarities between two images. For a grayscale image $f\in\ell^2(\bZ^2)$, the gradient map is defined by  
\begin{equation}
\label{eq:grad}
\operatorname{G_f}[x] = \sqrt{\left((g^{\text{hor}}*f)[x]\right)^2 + \left((g^{\text{ver}}*f)[x]\right)^2},
\end{equation}
where $g^{\text{hor}}$ and $g^{\text{ver}}$ denote horizontal and vertical gradient filters (e.g. Sobel or Scharr filters), and $*$ denotes the two-dimensional convolution operator. The method used in the implementation of the FSIM to compute the phase congruency map was developed by Peter Kovesi \cite{KovONLINE} and contains several non-trivial operations, such as adaptive soft thresholding. However, in its essence, the phase congruency map of a grayscale image $f$ is given by 
\begin{equation}
\label{eq:pc}
\operatorname{PC_{f}}[x] \approx \frac{\abs{\sum_n (g^\text{c}_n*f)[x]}}{\sum_n\abs{(g^\text{c}_n*f)[x]}},
\end{equation}

where $g^\text{c}_n$ denotes differently scaled and oriented complex-valued wavelet filters. The idea behind \eqref{eq:pc} is that if the obtained complex-valued wavelet coefficients have the same phase at a location $x$, taking the absolute value of the sum is the same as taking the sum of the absolute values. If this is the case, $\operatorname{PC_{f}}[x]$ will be close to or precisely $1$. 

To assess local similarities between two images with respect to the maps defined in \eqref{eq:grad} and \eqref{eq:pc}, the FSIM - like many other image quality metrics - uses a simple similarity measure for scalar values that already appeared in \cite{WBSS2004}, namely
\begin{equation}
\label{eq:scalarsim}
\Sim(a,b,C) = \frac{2ab + C}{a^2 + b^2 + C},
\end{equation}
with a constant $C > 0$. The graph of $\Sim(a,b,C)$ for values ranging from $0$ to $100$ and $C=30$ is shown in Figure~\ref{fig:scalarsim}. The local feature similarity map for two grayscale images $f_1,f_2\in\ell^2(\bZ^2)$ is defined by
\setlength{\arraycolsep}{0.0em}
\begin{equation}
\label{eq:localfs}
\operatorname{FS_{f_1,f_2}}[x]= \Sim\left(\operatorname{G_{f_1}}[x],\operatorname{G_{f_2}}[x],C_1\right)^\beta {\cdot} \Sim\left(\operatorname{PC_{f_1}}[x],\operatorname{PC_{f_2}}[x],C_2\right)^\gamma,
\end{equation}
\setlength{\arraycolsep}{5pt}
with constants $C_1,C_2>0$ and exponents $\beta,\gamma > 0$. Based on the assumption that the human visual system is especially sensitive towards structures at which the phases of the Fourier components are in congruency (see e.g. \cite{MRBO1986}), the phase congruency map is not only used in \eqref{eq:localfs} but also applied to determine the relative importance of different image areas with respect to human perception. Eventually, the feature similarity index is computed by taking the weighted mean of all local feature similarities, where the phase congruency map is used as a weight function, that is
\begin{equation}
\label{eq:fsim}
\operatorname{FSIM_{f_1,f_2}} = \frac{\sum_x \operatorname{FS_{f_1,f_2}}[x]  \cdot \operatorname{PC_{f_1,f_2}}[x]}{\sum_x \operatorname{PC_{f_1,f_2}}[x]},
\end{equation}
where
\begin{equation}
\operatorname{PC_{f_1,f_2}}[x] = \max\left(\operatorname{PC_{f_1}}[x],\operatorname{PC_{f_2}}[x]\right).
\end{equation}

The original publication of the FSIM proposes a generalization to color images defined in the YIQ color space, named FSIMC. In the YIQ space, the Y channel encodes luminance information, while the I and Q channels encode chromatic information. Color images defined in the RGB color space can easily be transformed to the YIQ space with a linear mapping, namely

\begin{equation}
\begin{bmatrix} f^{\text{Y}} \\ f^{\text{I}} \\ f^{\text{Q}} \end{bmatrix}
\approx
\begin{bmatrix} 0.299 & 0.587 & 0.114 \\ 0.596 & -0.274 & -0.322 \\ 0.211 & -0.523 & 0.312 \end{bmatrix}\cdot
\begin{bmatrix} f^{\text{R}} \\ f^{\text{G}} \\ f^{\text{B}} \end{bmatrix}.
\end{equation}

FSIMC simply incorporates the chroma channels I and Q into the local feature similarity measure \eqref{eq:localfs}. The gradient maps as well as the phase congruency maps are purely derived from the luminance channel Y in FSIMC and FSIM alike.

%Formally, FSIMC is given for color images $f_1,f_2$ by
%
%\begin{equation}
%\operatorname{FSIMC_{f_1,f_2}} = \frac{\sum_x \operatorname{FSC_{f_1,f_2}}[x]  \cdot \operatorname{PC_{f^{\text{Y}}_1,f^{\text{Y}}_2}}[x]}{\sum_x \operatorname{PC_{f^{\text{Y}}_1,f^{\text{Y}}_2}}[x]},
%\end{equation}
%
%where the local feature similarity measure is now defined as
%
%\setlength{\arraycolsep}{0.0em}
%\begin{eqnarray}
%\operatorname{FSC_{f_1,f_2}}[x]&{}={}&\Sim(\operatorname{G_{f^{\text{Y}}_1}}[x],\operatorname{G_{f^{\text{Y}}_2}}[x],C_1)^\alpha  \nonumber\\
%&&{\cdot}\:\Sim(\operatorname{PC_{f^{\text{Y}}_1}}[x],\operatorname{PC_{f^{\text{Y}}_2}}[x],C_2)^\beta \nonumber\\
%&&{\cdot}\:\Sim(f^{\text{I}}_1[x],f^{\text{I}}_2[x],C_3)^\gamma \nonumber\\
%&&{\cdot}\:\Sim(f^{\text{Q}}_1[x],f^{\text{Q}}_2[x],C_4)^\gamma,
%\end{eqnarray}
%\setlength{\arraycolsep}{5pt}
%with constants $C_1,C_2,C_3,C_4 > 0$ and exponents $\alpha,\beta,\gamma > 0$.

\section{The Haar Wavelet-Based Perceptual Similarity Index}
\label{sec:HaarPSI}

The basic idea of the HaarPSI is to construct feature maps in the spirit of \eqref{eq:grad} as well as a weight function similar to \eqref{eq:pc} by considering a single wavelet filterbank. The response of any high-frequency wavelet filter will look similar to the response yielded by a gradient filter like the Sobel operator. Furthermore, the phase congruency measure used as a weight function in the FSIM is computed directly from the output of a multi-scale complex-valued wavelet filterbank, as illustrated by Equation~\eqref{eq:pc}. This gives a strong intuition that it should be possible to define a similarity measure derived from the response of a single set of discrete wavelet filters that at least matches the performance of the FSIM on benchmark databases but requires significantly less computational effort.

\begin{figure}[!htb]
  \subfloat[Haar wavelet filters]{  \label{fig:haarfilters}
  \begin{tabular}[b]{cccc}   
    \includegraphics[width=0.08\textwidth]{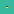}&
    \includegraphics[width=0.08\textwidth]{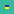}&   
    \includegraphics[width=0.08\textwidth]{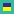}\\    \includegraphics[width=0.08\textwidth]{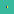}&
    \includegraphics[width=0.08\textwidth]{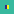}&   
    \includegraphics[width=0.08\textwidth]{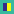}    
  \end{tabular}}\hfil
  \subfloat[$S(x,y,C)$]{  \label{fig:scalarsim}\includegraphics[width=0.25\textwidth]{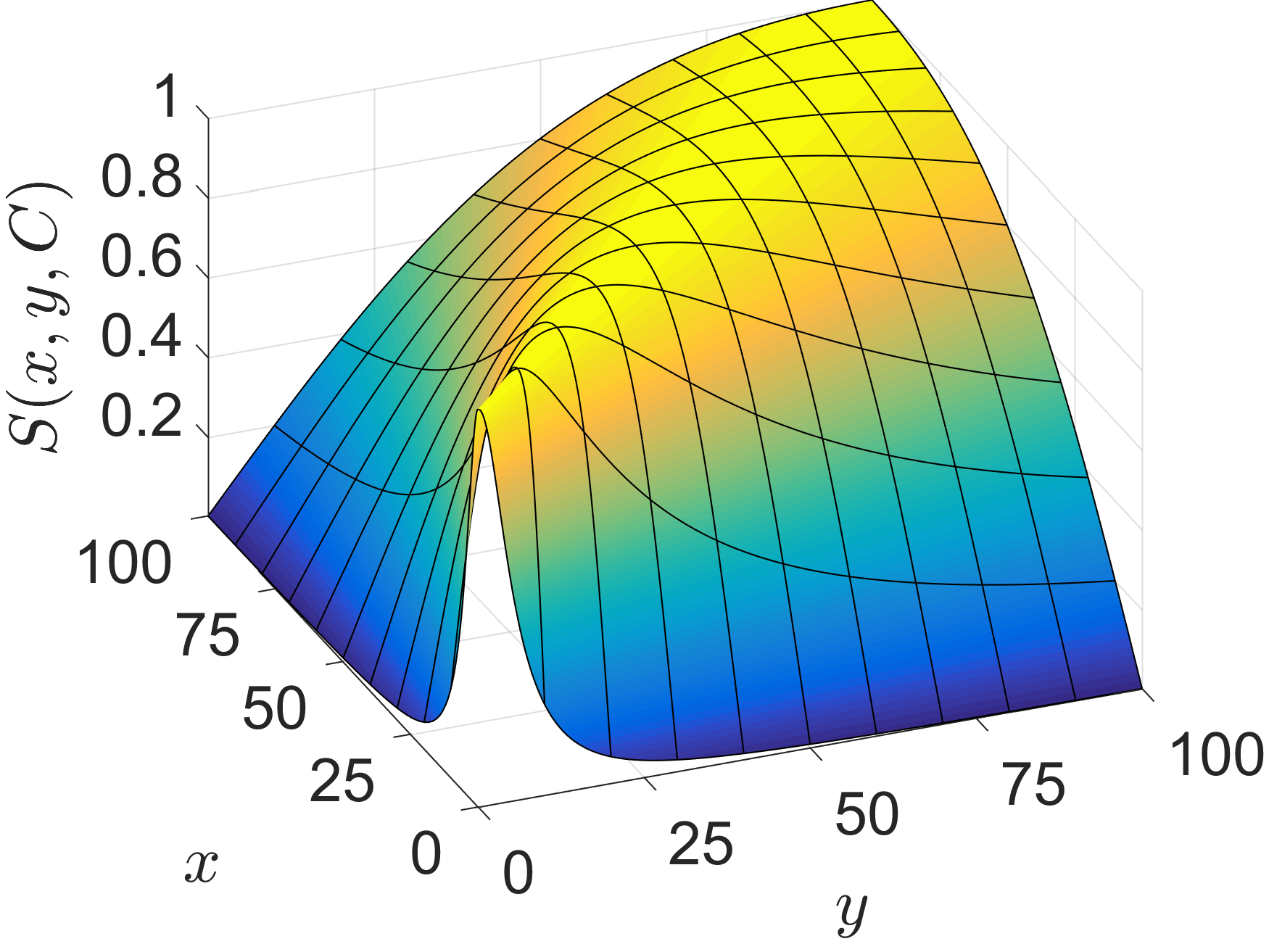}}\hfil
  \subfloat[$l_\alpha(x)$]{\label{fig:logistic}\includegraphics[width=0.25\textwidth]{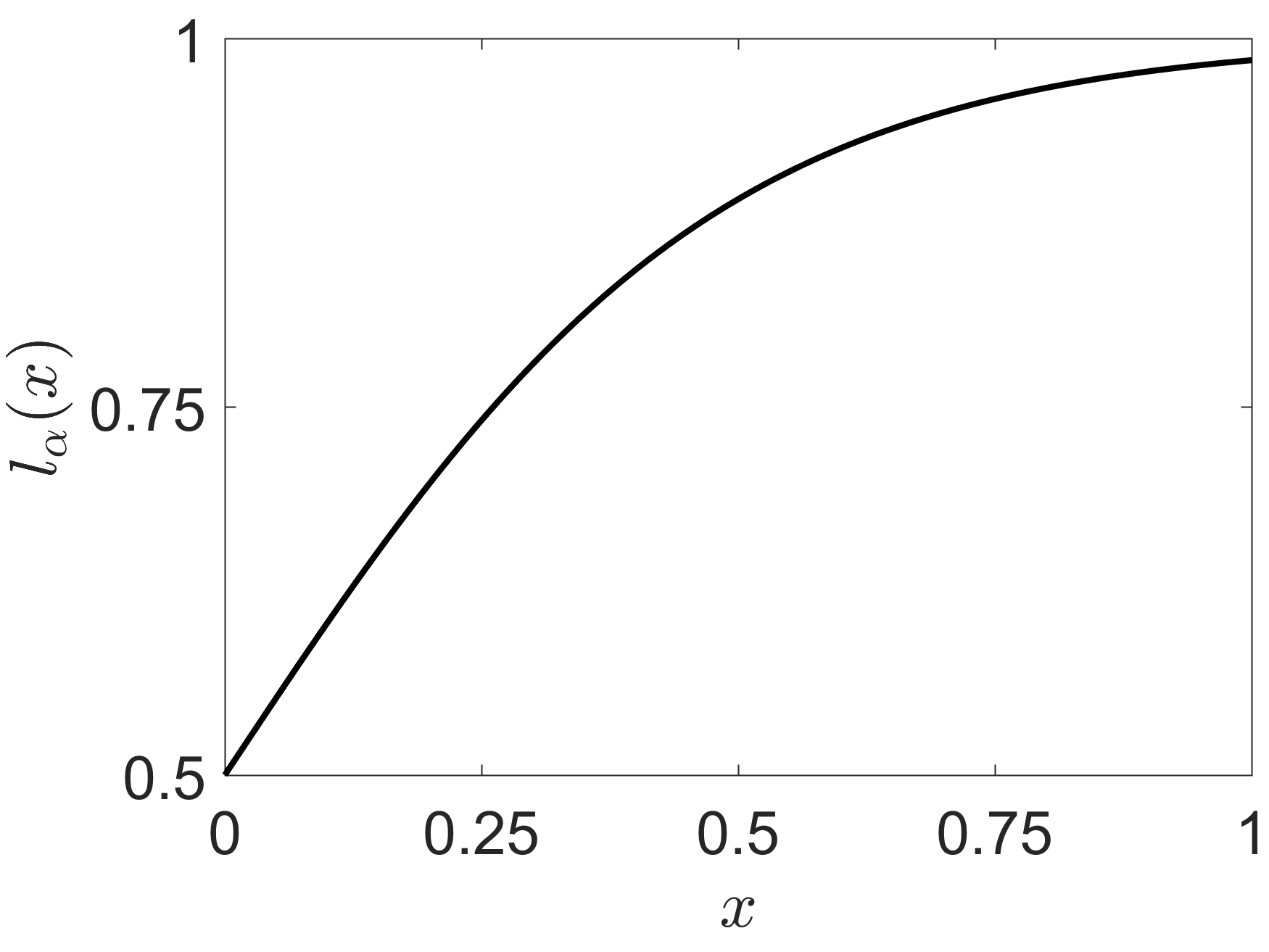}}
  \caption{(a) The six Haar wavelet filters whose responses build the core of the HaarPSI. (b)  The function $S(x,y,C)$ for $C=30$. (c) The logistic function $l_\alpha(x)$ for $\alpha = 4.2$.}
\end{figure}

The wavelet chosen for this endeavor is the so-called Haar wavelet, which was already proposed in 1910 by Alfred Haar \cite{Haar1910} and is arguably the simplest and computationally most efficient wavelet there is. The one-dimensional Haar filters are given by
\begin{equation}
\label{eq:haarfilters}
h^{\text{1D}}_1 = \frac{1}{\sqrt{2}}\cdot[1,1] \text{ and } g^{\text{1D}}_1 = \frac{1}{\sqrt{2}}\cdot[-1,1],
\end{equation}
where $h^{\text{1D}}_1$ denotes the low-pass scaling filter and $g^{\text{1D}}_1$ the corresponding high-pass wavelet filter.
For any scale $j\in\bN$, we can construct two-dimensional Haar filters by setting
\begin{align*}
g^{\text{(1)}}_j &= g^{\text{1D}}_j \otimes h^{\text{1D}}_j,\\
g^{\text{(2)}}_j &= h^{\text{1D}}_j \otimes g^{\text{1D}}_j,
\end{align*}
where $\otimes$ denotes the outer product and the one-dimensional filters $h^{\text{1D}}_j$ and $g^{\text{1D}}_j$ are given for $j>1$ by
\begin{align*}
g^{\text{1D}}_j &= h^{\text{1D}}_{1}*(g^{\text{1D}}_{j-1})_{\uparrow 2}, \\
h^{\text{1D}}_j &= h^{\text{1D}}_{1}*(h^{\text{1D}}_{j-1})_{\uparrow 2},
\end{align*}
where $\uparrow2$ is the dyadic upsampling operator, and $*$ denotes the one-dimensional convolution operator. Note that $g^{\text{(1)}}_j$ responds to horizontal structures, while  $g^{\text{(2)}}_j$ picks up vertical structures. The six Haar filters used to define the HaarPSI are shown in Figure~\ref{fig:haarfilters}.

\setlength{\tabcolsep}{1mm}

The local similarity map $\operatorname{FS_{f_1,f_2}}$ multiplicatively combines gradient-based and phase congruency-based similarities whose contributions are weighted by the exponents $\alpha,\beta>0$. The HaarPSI does not consider different types of similarities. However, to correctly predict the perceptual similarity experienced by human viewers, it can be useful to apply an additional non-linear mapping to the local similarities obtained from high-frequency Haar wavelet filter responses. This non-linearity is chosen to be the logistic function, which is widely used as an activation function in neural networks {for modeling thresholding in biological neurons}  and {is} given for a parameter $\alpha >0$ as
\begin{equation}
 \label{eq:logistic}
 l_\alpha(x) = \frac{1}{1 + e^{-\alpha x}}.
\end{equation}

\begin{figure}[!htb]
  \centering
  \begin{minipage}{0.28\textwidth}
	\subfloat[reference $f_1$]{\includegraphics[width=1\textwidth]{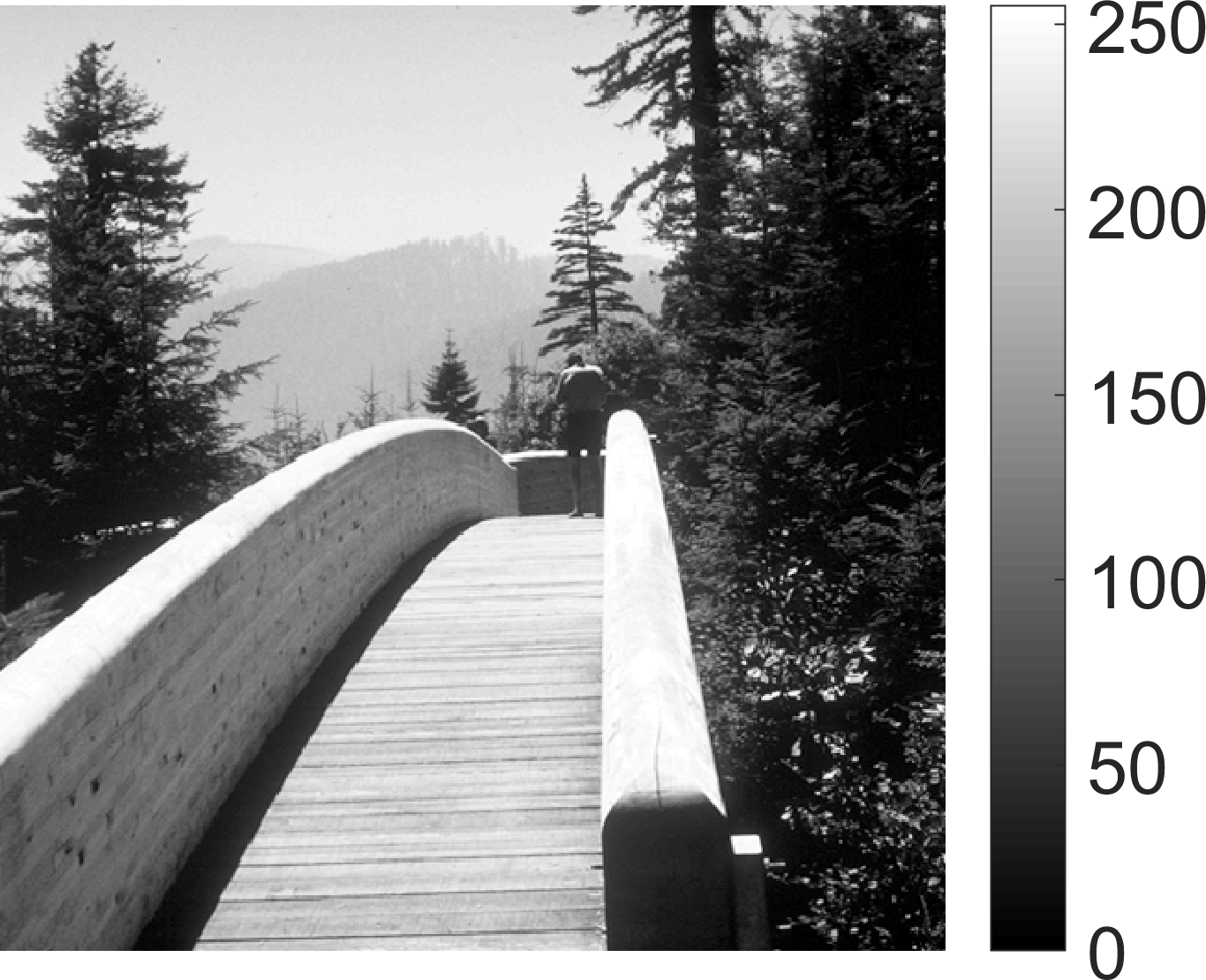}}\\
	\subfloat[distorted $f_2$]{\includegraphics[width=1\textwidth]{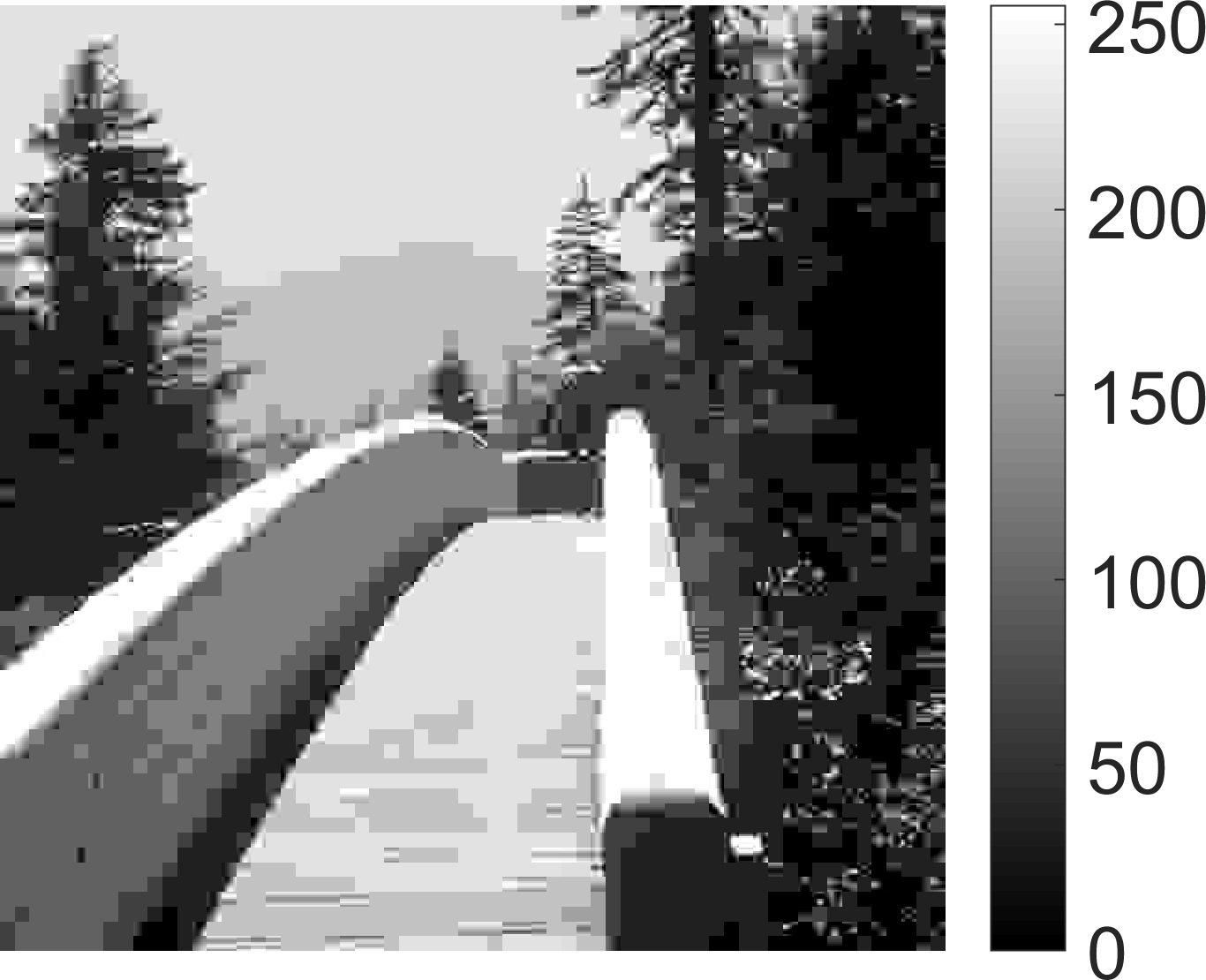}}
  \end{minipage}\hfil  
  \begin{minipage}{0.28\textwidth}  
  \subfloat[$\operatorname{HS^\text{(1)}_{f_1,f_2}}$]{\includegraphics[width=1\textwidth]{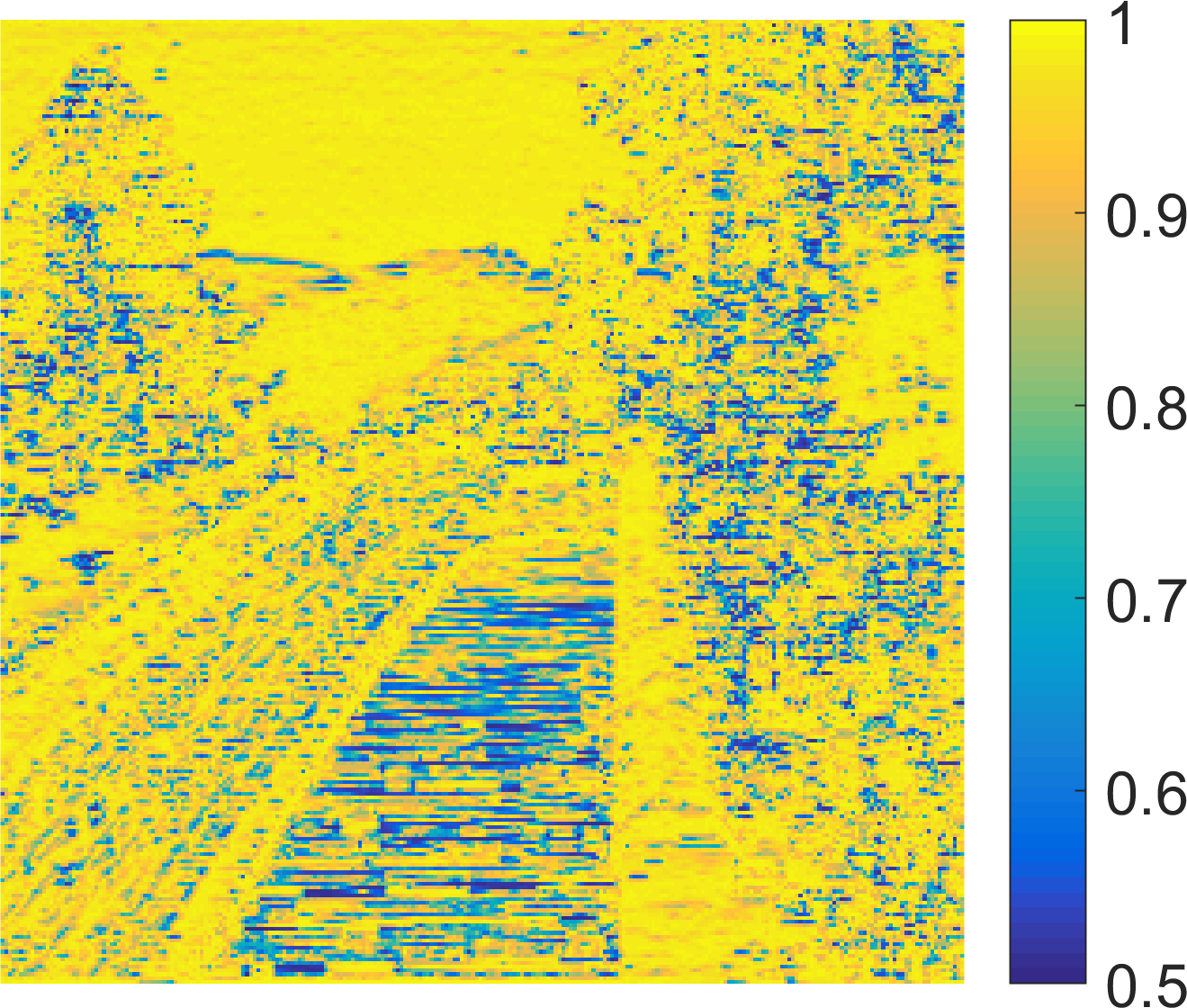}}\\
  \subfloat[$\operatorname{HS^\text{(2)}_{f_1,f_2}}$]{\includegraphics[width=1\textwidth]{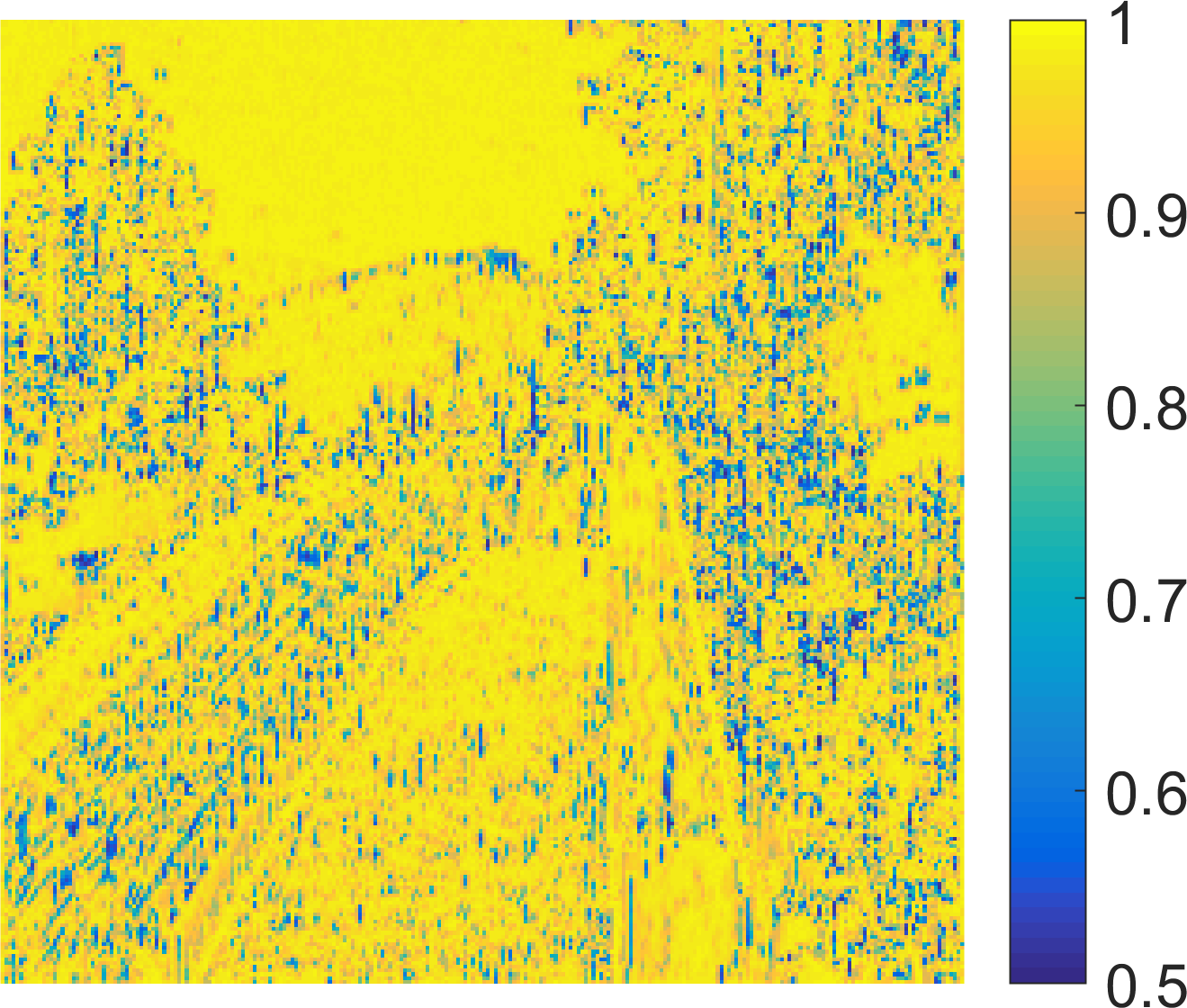}}
\end{minipage}\hfil  
\begin{minipage}{0.28\textwidth}
\subfloat[$\operatorname{W^\text{(1)}_{f_1,f_2}}$]{\includegraphics[width=1\textwidth]{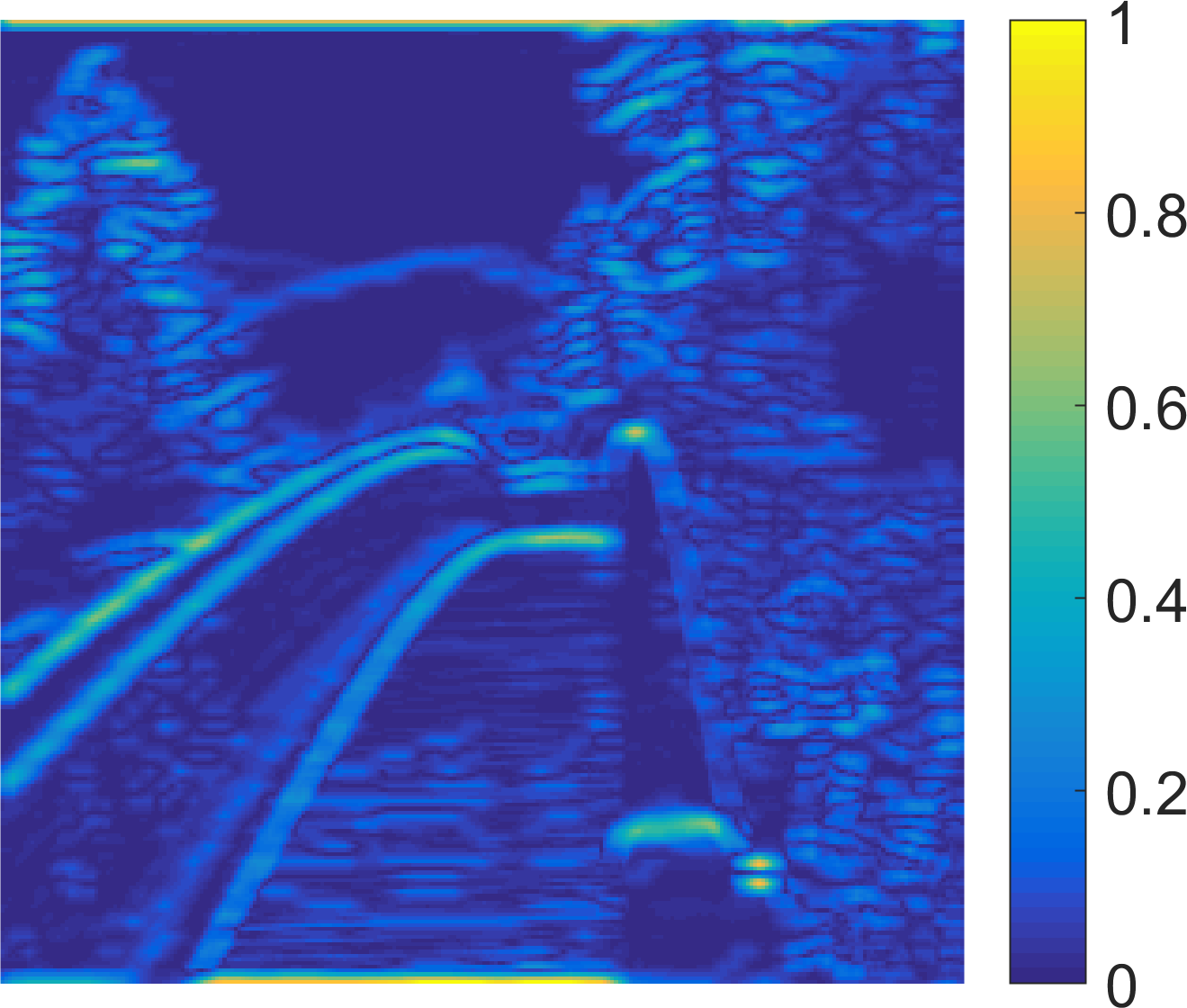}}\\
\subfloat[$\operatorname{W^\text{(2)}_{f_1,f_2}}$]{\includegraphics[width=1\textwidth]{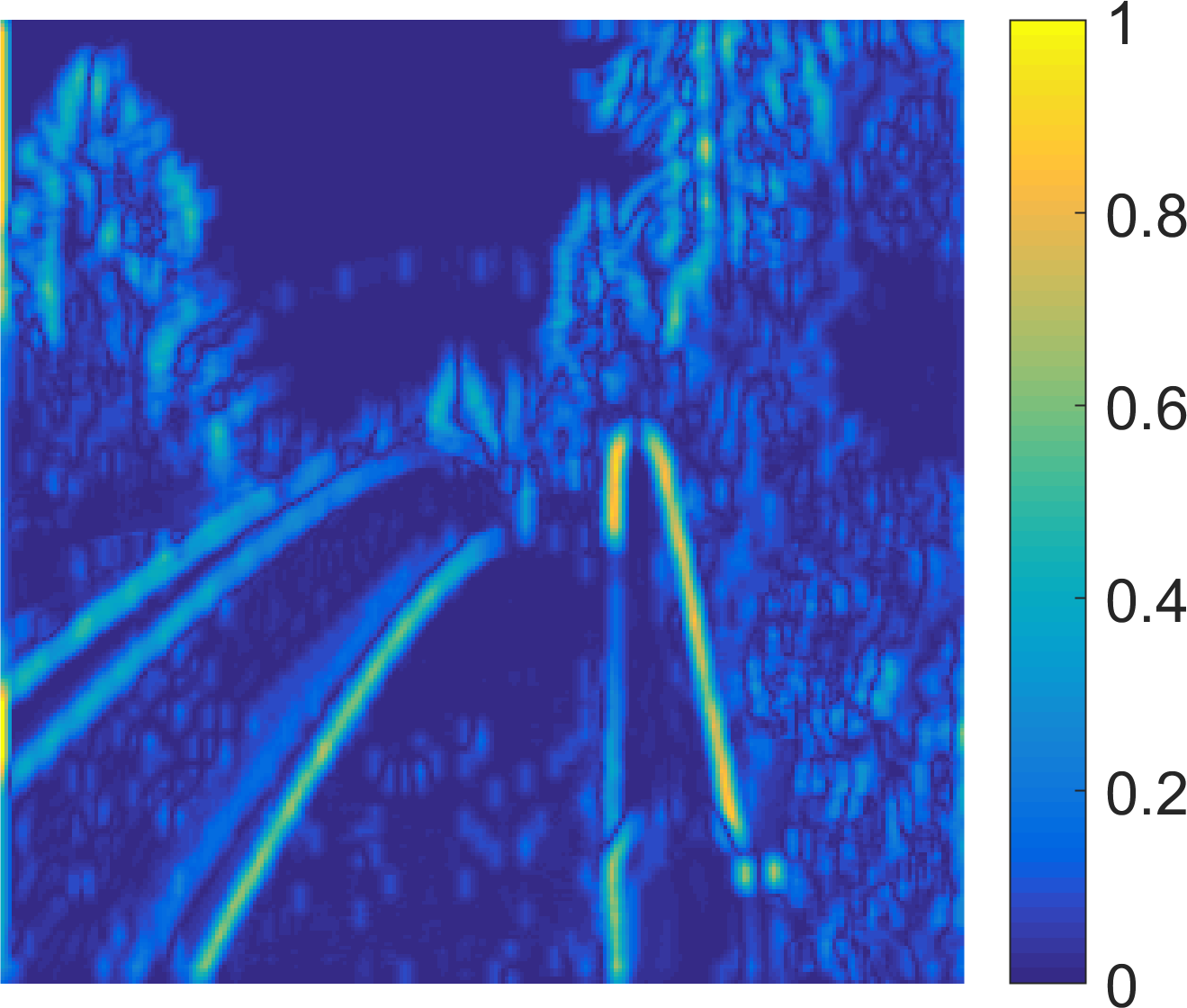}}
\end{minipage}  
  \caption{(a) An undistorted reference image. (b) The reference image distorted by the JPEG compression algorithm. (c) The horizontal local similarity map $\operatorname{HS^\text{(1)}_{f_1,f_2}}$. (d) The vertical local similarity map $\operatorname{HS^\text{(2)}_{f_1,f_2}}$. (e) The (normalized) horizontal weight function $\operatorname{W^\text{(1)}_{f_1,f_2}}$. (f) The (normalized) vertical weight function $\operatorname{W^\text{(2)}_{f_1,f_2}}$. The images (a) and (b) are part of the CSIQ database \cite{LaCh2010}.}
  \label{fig:haarmaps}
\end{figure}

For two grayscale images $f_1,f_2\in\ell^2(\bZ^2)$, the local similarity measure used to compute the HaarPSI is based on the first two stages of a two-dimensional discrete Haar wavelet transform and given by
\begin{equation}
\label{eq:localhs}
\operatorname{HS^\text{(k)}_{f_1,f_2}}[x]=l_\alpha\left(\frac{1}{2}\sum_{j= 1}^2\Sim\left(\abs{(g^\text{(k)}_j*f_1)[x]},\abs{(g^\text{(k)}_j*f_2)[x]},C\right)\right), 
\end{equation}
where $C > 0$, $k\in\{1,2\}$ selects either horizontal or vertical Haar wavelet filters, $\Sim$ denotes the similarity measure \eqref{eq:scalarsim}, and $*$ is the two-dimensional convolution operator. The local similarity measure $\operatorname{HS^\text{(k)}_{f_1,f_2}}$ can be seen as an analog to $\operatorname{FS_{f_1,f_2}}$. However, $\operatorname{HS^\text{(k)}_{f_1,f_2}}$ does not mix different different concepts like gradients and phase congruency and is computed straightforwardly on the responses of two high-frequency discrete Haar wavelet filters. A visualization of the local similarity map $\operatorname{HS^\text{(k)}_{f_1,f_2}}$ is shown in Figure~\ref{fig:haarmaps}.

Analogous to the phase congruency map $\operatorname{PC_{f}}$ in the definition of the FSIM, the HaarPSI considers a weight map which is derived from the response of a single low-frequency Haar wavelet filter:
\begin{equation}
\label{eq:haarweight}
\operatorname{W^\text{(k)}_{f}}[x] = \abs{(g^{\text{(k)}}_3*f)[x]},
\end{equation}
where $k\in\{1,2\}$ again differentiates between horizontal and vertical filters. Figure~\ref{fig:haarmaps} shows an example of the weight map $\operatorname{W^\text{(k)}_{f}}$ computed from a natural image.

The Haar-wavelet based perceptually similarity index for two grayscale images $f_1,f_2$ is eventually given as the weighted average of the local similarity map $\operatorname{HS^\text{(k)}_{f_1,f_2}}$ , that is, 
\begin{equation}
\label{eq:haarpsi}
\operatorname{HaarPSI_{f_1,f_2}} = l_\alpha^{-1}\left(\frac{\sum\limits_x \sum\limits_{k=1}^2\operatorname{HS^\text{(k)}_{f_1,f_2}}[x]  \cdot \operatorname{W^\text{(k)}_{f_1,f_2}}[x]}{\sum\limits_x \sum\limits_{k=1}^2\operatorname{W^\text{(k)}_{f_1,f_2}}[x]}\right)^2, 
\end{equation}
with
\begin{equation}
\operatorname{W^\text{(k)}_{f_1,f_2}}[x] = \max(\operatorname{W^\text{(k)}_{f_1}}[x],\operatorname{W^\text{(k)}_{f_2}}[x])
\end{equation}
for $k\in \{1,2\}$.
The function $l_\alpha^{-1}(\cdot)$ maps the weighted average from the interval $[\frac{1}{2},l_\alpha(1)]$ back to $[0,1]$. Applying $(\cdot)^2$ further spreads the HaarPSI in the unit interval and helps to linearize the relationship between the HaarPSI and human opinion scores. In particular, this procedure aims to increase the readability of the HaarPSI in the sense that a single value should be 'meaningful on its own' and not only relative to other HaarPSI values. Please note that, due to the monotonicity of the logistic function, applying $l_\alpha^{-1}(\cdot)^2$ cannot improve or worsen the rank order-based correlations with human opinion scores reported in Section~\ref{sec:results}.

\

Analogous to the FSIM, the HaarPSI can be extended to color images in the YIQ color space by considering a third local similarity map based on the chroma channels I and Q. The map $\operatorname{HS^\text{(3)}_{f_1,f_2}}$ is computed analogous to \eqref{eq:localhs} by averaging local similarities obtained from comparing $f^{\text{I}}_1$ with $f^{\text{I}}_2$ and $f^{\text{Q}}_1$ with $f^{\text{Q}}_2$. In contrast to $\operatorname{HS^\text{(1)}_{f_1,f_2}}$ and $\operatorname{HS^\text{(2)}_{f_1,f_2}}$, the chromatic information used for $\operatorname{HS^\text{(3)}_{f_1,f_2}}$ is not based on orientation sensitive filters. The corresponding weight map $\operatorname{W^\text{(3)}_{f^\text{Y}_1,f^\text{Y}_2}}$ is thus also computed by averaging $\operatorname{W^\text{(1)}_{f^\text{Y}_1,f^\text{Y}_2}}$ and $\operatorname{W^\text{(2)}_{f^\text{Y}_1,f^\text{Y}_2}}$. Formally, the generalization of the HaarPSI to color images is given by
\begin{equation}
\label{eq:haarpsic}
\operatorname{HaarPSIC_{f_1,f_2}} = l_\alpha^{-1}\left(\frac{\sum\limits_x \sum\limits_{k=1}^3\operatorname{HS^\text{(k)}_{f_1,f_2}}[x] \cdot \operatorname{W^\text{(k)}_{f^\text{Y}_1,f^\text{Y}_2}}[x]}{\sum\limits_x \sum\limits_{k=1}^3\operatorname{W^\text{(k)}_{f^\text{Y}_1,f^\text{Y}_2}}[x]}\right)^2, 
\end{equation}
with $\operatorname{HS^\text{(1)}_{f_1,f_2}}$ and $\operatorname{HS^\text{(2)}_{f_1,f_2}}$ defined as in \eqref{eq:localhs},
\begin{equation}
\label{eq:localhs3}
\resizebox{.9\hsize}{!}{$\operatorname{HS^\text{(3)}_{f_1,f_2}}[x]=l_\alpha\left(\frac{1}{2}\left(\Sim\left(\abs{(m*f^{\text{I}}_1)[x]},\abs{(m*f^{\text{I}}_2)[x]},C\right) + \Sim(\abs{(m*f^{\text{Q}}_1)[x]},\abs{(m*f^{\text{Q}}_2)[x]},C)\right)\right)$,}
\end{equation}
\setlength{\arraycolsep}{5pt}
with a $2\times2$ mean filter $m$ and
\begin{equation}
\operatorname{W^\text{(3)}_{f^\text{Y}_1,f^\text{Y}_2}}[x] = \frac{1}{2}\left(\operatorname{W^\text{(1)}_{f^\text{Y}_1,f^\text{Y}_2}}[x]+\operatorname{W^\text{(2)}_{f^\text{Y}_1,f^\text{Y}_2}}[x]\right).
\end{equation}

\subsection{Parameter Selection}

\begin{figure}
	\setlength{\tabcolsep}{3mm}
	\subfloat[]{\begin{scriptsize}
			\begin{threeparttable}[b]  
				\begin{tabular}[b]{*{5}{c}}\toprule[0.5mm]
					&  & All databases & TID only & LIVE \& CSIQ only\\
					& $C$ & 30 & 30  & 20  \\
					& $\alpha$ & 4.2 & 4.2 & 5.8\\[0.1cm]
					\multicolumn{5}{c}{Spearman Rank Order Correlations (SROCC)}\\[0.1cm]
	      & LIVE & \textbf{0.9683} & \textbf{0.9683} & 0.9677\\
	      & TID2008 & \textbf{0.9097} & \textbf{0.9097} & 0.9031\\
	      & TID2013 & \textbf{0.8732} & \textbf{0.8732} & 0.8651\\
	      & CSIQ & 0.9604 & 0.9604 & \textbf{0.9625}\\
					\hline
				\end{tabular}
				\begin{tablenotes}
					\item The highest correlation in each row is written in boldface.
				\end{tablenotes}
			\end{threeparttable}
		\end{scriptsize}}
		\hfill
		\subfloat[]{\includegraphics[width = 0.4\textwidth]{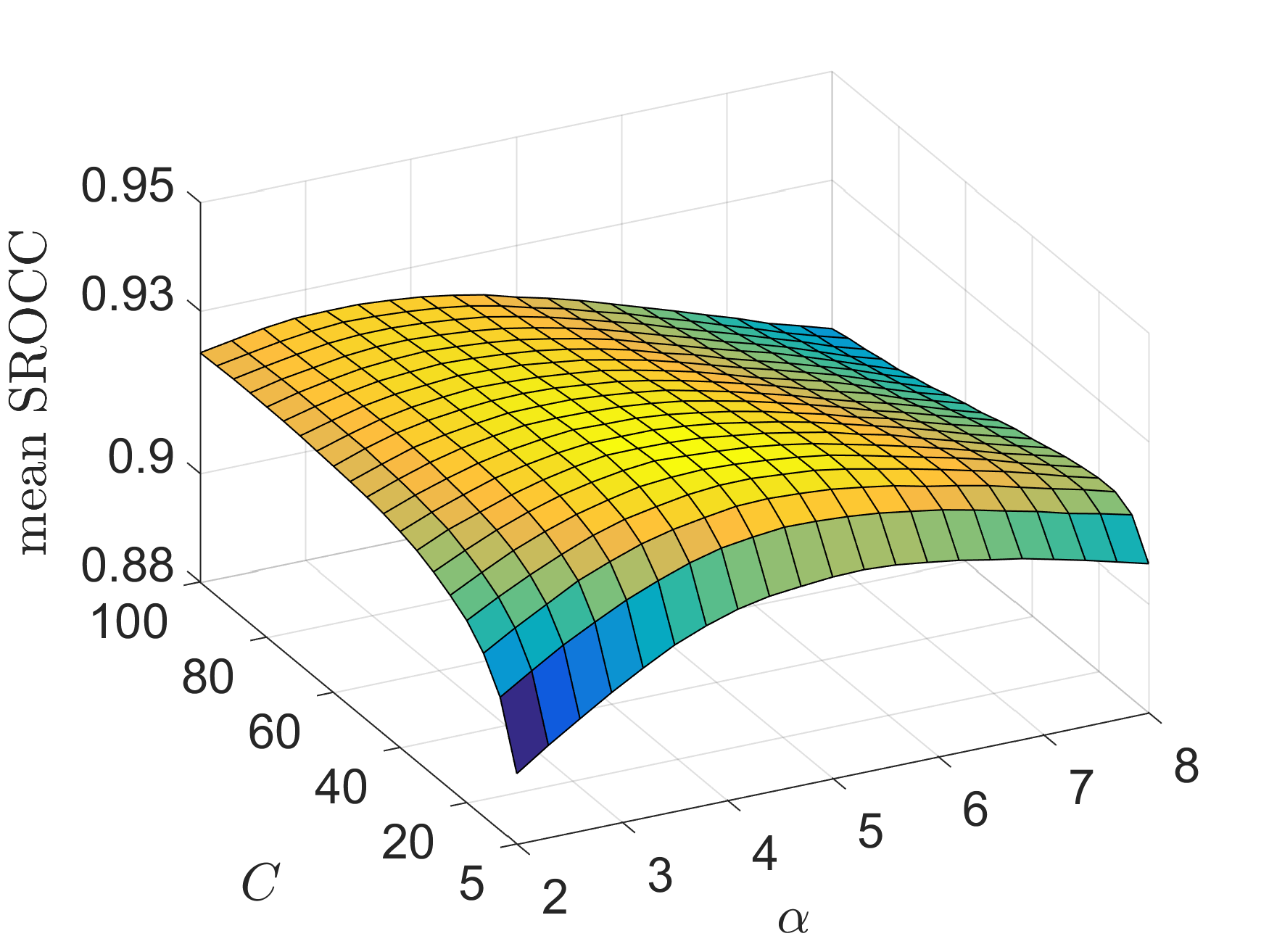}}
		\caption{(a) Values for the parameters $C$ and $\alpha$ which maximize the mean SROCC with respect to randomly selected subsets of the considered databases. The values in the first column were obtained by including all four databases in the optimization procedure. For the results depicted in columns 2 and 3, the optimization was restricted to the TID 2008 \& TID 2013 respectively the LIVE \& CSIQ databases. The SROCC values shown in the last four rows are with respect to the full databases. (b) The mean SROCC with respect to the subsets of all four databases plotted as a function of the parameters $C$ and $\alpha$.}
		\label{fig:optimization}
	\end{figure}

The HaarPSI as well as the HaarPSIC require only two parameters to be selected, namely $C$ and $\alpha$. Both parameters were optimized on randomly chosen subsets of four large publicly available databases, where each subset was a quarter the size of the original database. Each of the databases, which will be described in more detail in Section~\ref{sec:results}, contains large numbers of differently distorted images and their corresponding MOS values.
The parameters $C$ and $\alpha$ were selected to maximize the mean of the four Spearman rank order correlation coefficients (SROCC) obtained from comparing HaarPSIC and MOS values from subsets of the TID 2008 \cite{PLZECB2009}, TID 2013  \cite{Ponomarenko2015}, LIVE \cite{SWCBOnline} and CSIQ \cite{LaCh2010} image databases. The optimization was carried out in two steps. First, a grid search was performed in which the parameter $C$ took values in the interval $[5,100]$ and $\alpha$ in the range between $2$ and $8$. The best $(C,\alpha)$ pair was then used as the initial value of the Nelder-Mead algorithm. The thus refined parameters were eventually rounded to the nearest integer in the case of $C$ and to the nearest tenth in the case of $\alpha$. This procedure resulted in the choices of $C=30$ and $\alpha = 4.2$.
%cross validation procedure
To verify the generality of the HaarPSI, the same optimization procedure was repeated once only considering the TID 2008 and TID 2013 databases and once restricted to the LIVE and the CSIQ image databases. The results of all three optimizations are compiled in Figure~\ref{fig:optimization}.

\section{Experimental Results}
\label{sec:results}
% different databases
The consistency of the HaarPSI with the human perception of image quality was
evaluated and compared with most of the image quality metrics discussed in
Section~\ref{sec:intro} on four large publicly available benchmark databases
of quality-annotated images. Those databases differ in the number of reference images, the number of distortion magnitudes and types, the number of observers, the level of control of the viewing conditions, and the stimulus presentation procedure.

% LIVE
The LIVE database \cite{SWCBOnline} contains 29 reference color images and 779 distorted images that were perturbed by JPEG compression, JPEG 2000 compression, additive Gaussian white
noise, Gaussian blurring as well as JPEG 2000 compressed images that have been transmitted over a simulated Rayleigh fading channel. Each distortion is introduced at five to six different levels of magnitude. On average, about 23 subjects evaluated the quality of each image with respect to the reference image. The viewing conditions were fairly controlled for in terms of viewing distance. Ratings were collected in a double stimulus manner.

% TID 2008
The TID 2008 database \cite{PLZECB2009} comprises 25 colored reference images and 1700 degraded images, that had been subject to a wide range of distortions, including various types of noise, blur, JPEG and JPEG 2000 compression, transmission errors, local image distortions, as well as luminance and contrast changes. Subjective ratings were gathered by comparisons. The results from several viewing conditions of experiments in three different labs and on the internet were averaged. TID 2008 was later extended to TID 2013 \cite{Ponomarenko2015}, which added new types of distortions, which are mostly of a chromatic nature. In total, TID 2013 contains 3000 differently distorted images.

% CSIQ
The CSIQ database \cite{LaCh2010} is based on 30 reference color images and contains 866 distorted images. Six different types of distortions (JPEG compression, JPEG 2000 compression, global contrast decrements, additive pink Gaussian noise, and Gaussian blurring) at four to five different degradation magnitudes were applied to the reference images. The viewing distance was controlled. Images were presented on a monitor array and subjects were asked to place all distorted versions of one reference image according to its perceived quality.

% Why SROCC
The main goal of most computational image similarity measures is to yield a monotonic relationship with human mean opinion scores across different databases and distortion types. To ensure a fair evaluation, different computational measures are typically compared with respect to rank order-based correlations or after performing nonlinear regression. Throughout the numerical evaluation of the HaarPSI, we apply the rank order-based SROCC to measure correlations between human mean opinion scores and different computational similarity and distortion indexes. We also considered applying Kendall's $\tau$ and the Pearson product-moment correlation after performing a four parameter logistic regression as alternatives for the SROCC. We found that these correlation coefficients essentially duplicate the results reported in this section. The corresponding versions of Tables~\ref{tab:sroccdatabases}~and~\ref{tab:details} were thus not included here but can be found at \url{www.haarpsi.org}.

% significance test
 Following the ITU guidelines for evaluating quality prediction models \cite{ITUTP1401}, we also tested the statistical significance of the results reported in this section. Correlation coefficients for which the $H_0$ hypothesis that they are not significantly different than the respective HaarPSI correlation can be refuted with $p < 0.05$ are highlighted in color in Tables~\ref{tab:sroccdatabases},~\ref{tab:details}~and~\ref{tab:otherwavelets}. In accordance with \cite{fieller1957tests}, the variance of the z-transforms were approximated by $1.06/(N - 3)$, where $N$ denotes the degrees of freedom (i.e. the number of samples in the considered database or distortion specific subset).

\setlength{\tabcolsep}{2mm}
\begin{table}[!htb]
	\centering
	\caption{Spearman Rank Order Correlations of IQA Metrics With Human Mean Opinion Scores}
	\label{tab:sroccdatabases}	
	\begin{threeparttable}
		\begin{scriptsize}
			\begin{tabular}{*{12}{c}}
				\toprule[0.5mm]
				\multicolumn{12}{c}{Grayscale Images}\\[0.1cm]
				& & PSNR & VIF & SSIM & MS-SSIM & GSM & MAD & SR-SIM & FSIM & VSI & HaarPSI\\
				& LIVE & \cellcolor{green!25}0.8756 & 0.9636 & \cellcolor{green!25}0.9479 & \cellcolor{green!25}0.9513 & \cellcolor{green!25}0.9561 & 0.9672 & \cellcolor{green!25}0.9619 & 0.9634 & \cellcolor{green!25}0.9534 & \textbf{0.9690}\\
				& TID2008 & \cellcolor{green!25}0.5531 & \cellcolor{green!25}0.7491 & \cellcolor{green!25}0.7749 & \cellcolor{green!25}0.8542 & \cellcolor{green!25}0.8504 & \cellcolor{green!25}0.8340 & 0.8913 & \cellcolor{green!25}0.8804 & \cellcolor{green!25}0.8830 & \textbf{0.9043}\\
				& TID2013 & \cellcolor{green!25}0.6394 & \cellcolor{green!25}0.6769 & \cellcolor{green!25}0.7417 & \cellcolor{green!25}0.7859 & 0.7946 & \cellcolor{green!25}0.7807 & 0.8075 & 0.8022 & 0.8048 & \textbf{0.8094}\\
				& CSIQ & \cellcolor{green!25}0.8058 & \cellcolor{green!25}0.9195 & \cellcolor{green!25}0.8756 & \cellcolor{green!25}0.9133 & \cellcolor{green!25}0.9108 & 0.9466 & \cellcolor{green!25}0.9319 & \cellcolor{green!25}0.9242 & \cellcolor{green!25}0.9372 & \textbf{0.9546}\\
				\\
				\multicolumn{12}{c}{Color Images}\\[0.1cm]
				& & PSNR & VIF & SSIM & MS-SSIM & GSM & MAD & SR-SIM & FSIM & VSI & HaarPSI\\
				& LIVE & \cellcolor{green!25}0.8756 & 0.9636 & \cellcolor{green!25}0.9479 & \cellcolor{green!25}0.9513 & \cellcolor{green!25}0.9561 & 0.9672 & 0.9619 & 0.9645 & \cellcolor{green!25}0.9524 & \textbf{0.9683}\\
				& TID2008 & \cellcolor{green!25}0.5531 & \cellcolor{green!25}0.7491 & \cellcolor{green!25}0.7749 & \cellcolor{green!25}0.8542 & \cellcolor{green!25}0.8504 & \cellcolor{green!25}0.8340 & \cellcolor{green!25}0.8913 & \cellcolor{green!25}0.8840 & 0.8979 & \textbf{0.9097}\\
				& TID2013 & \cellcolor{green!25}0.6394 & \cellcolor{green!25}0.6769 & \cellcolor{green!25}0.7417 & \cellcolor{green!25}0.7859 & \cellcolor{green!25}0.7946 & \cellcolor{green!25}0.7807 & \cellcolor{green!25}0.8075 & \cellcolor{green!25}0.8510 & \cellcolor{red!25}\textbf{0.8965} & 0.8732\\
				& CSIQ & \cellcolor{green!25}0.8058 & \cellcolor{green!25}0.9195 & \cellcolor{green!25}0.8756 & \cellcolor{green!25}0.9133 & \cellcolor{green!25}0.9108 & \cellcolor{green!25}0.9466 & \cellcolor{green!25}0.9319 & \cellcolor{green!25}0.9310 & \cellcolor{green!25}0.9423 & \textbf{0.9604}\\\midrule[0.5mm]
			\end{tabular}
			\begin{tablenotes}
				\item \colorbox{green!25}{Lower correlation than HaarPSI. The difference is statistically significant with $p < 0.05$.}
				\item \colorbox{red!25}{Higher correlation than HaarPSI. The difference is statistically significant with $p < 0.05$.}
				\item The highest correlation in each row is written in \textbf{boldface}.
			\end{tablenotes}
	\end{scriptsize}
\end{threeparttable}
\end{table}

% color grayscale
The four databases used in the numerical evaluation only contain color images. However, out of the metrics considered in our experiments, only the FSIM and the HaarPSI are defined for both grayscale and color images, while the visual saliency-based index (VSI) was specifically designed for color images. All other similarity measures considered in our experiments only accept grayscale images as input or perform an RGB to grayscale conversion as a first processing step. To reflect these differing designs, all methods were tested on all databases once with the original color images and once with grayscale conversions obtained from the Matlab \textit{rgb2gray} function. To obtain the VSI for pairs of grayscale images, corresponding RGB images were created by setting the values for all three color channels to the values of the given grayscale channel. The correlation coefficients of all ten considered similarity measures with the human mean opinion scores for the LIVE image database, TID 2008, TID 2013 and the CSIQ database are compiled in Table~\ref{tab:sroccdatabases}.

%overall results and time
Table~\ref{tab:overallperf} provides a quick impression of the overall performance of each metric. It depicts the average SROCC of each metric with respect to all four databases as well as the mean execution time in milliseconds. The average execution time was measured on a Intel Core i7-4790 CPU clocked at $3.60$ GHz. To measure the execution time, each quality measure was computed ten times for ten different pairs of randomly generated $512\times512$ pixel images. All computations and measurements were carried out in Matlab using implementations made freely available by the respective authors. Note that due to an additional conversion step, metrics that are only defined for grayscale images can have slightly higher execution times when evaluated on color images.

\begin{table}[!htb]
  \centering
  \caption{Mean SROCC and Execution Time} 
  \label{tab:overallperf}
\begin{small}
  \begin{tabular}{*{6}{c}}
    \toprule[0.5mm] &  & \multicolumn{2}{c}{Color Images} & \multicolumn{2}{c}{Grayscale Images}\\
    &  & SROCC & Time (ms) & SROCC & Time (ms)\\
& HaarPSI & 0.9279 & 24 & 0.9093 & 10\\
& VSI & 0.9223 & 79 & 0.8946 & 80\\
& FSIM & 0.9076 & 142 & 0.8925 & 121\\
& SRSIM & 0.8982 & 10 & 0.8982 & 10\\
& MAD & 0.8821 & 892 & 0.8821 & 891\\
& GSM & 0.8780 & 8 & 0.8780 & 7\\
& MSSSIM & 0.8762 & 30 & 0.8762 & 24\\
& SSIM & 0.8350 & 6 & 0.8350 & 5\\
& VIF & 0.8273 & 459 & 0.8273 & 453\\
& PSNR & 0.7185 & 2 & 0.7185 & 1\\
    \midrule[0.5mm]
  \end{tabular}
\end{small}
\end{table}

%detailed results.
A high correlation with the mean opinion scores annotated to the distorted images of a large database containing many different types and degrees of distortions is arguably the best indicator of an image quality measure's consistency with human perception. However, for certain applications like compression or denoising, it could be more important to know if an image quality metric has a high correlation with the human experience \textit{within} a single distortion class. Table~\ref{tab:details} depicts the SROC coefficients for all image quality metrics when only subsets of databases containing specific distortions like Gaussian blur or JPEG transmission errors are considered. 

\begin{figure}[!htb]
	\begin{minipage}{0.35\linewidth}
		\includegraphics[width = 1\textwidth]{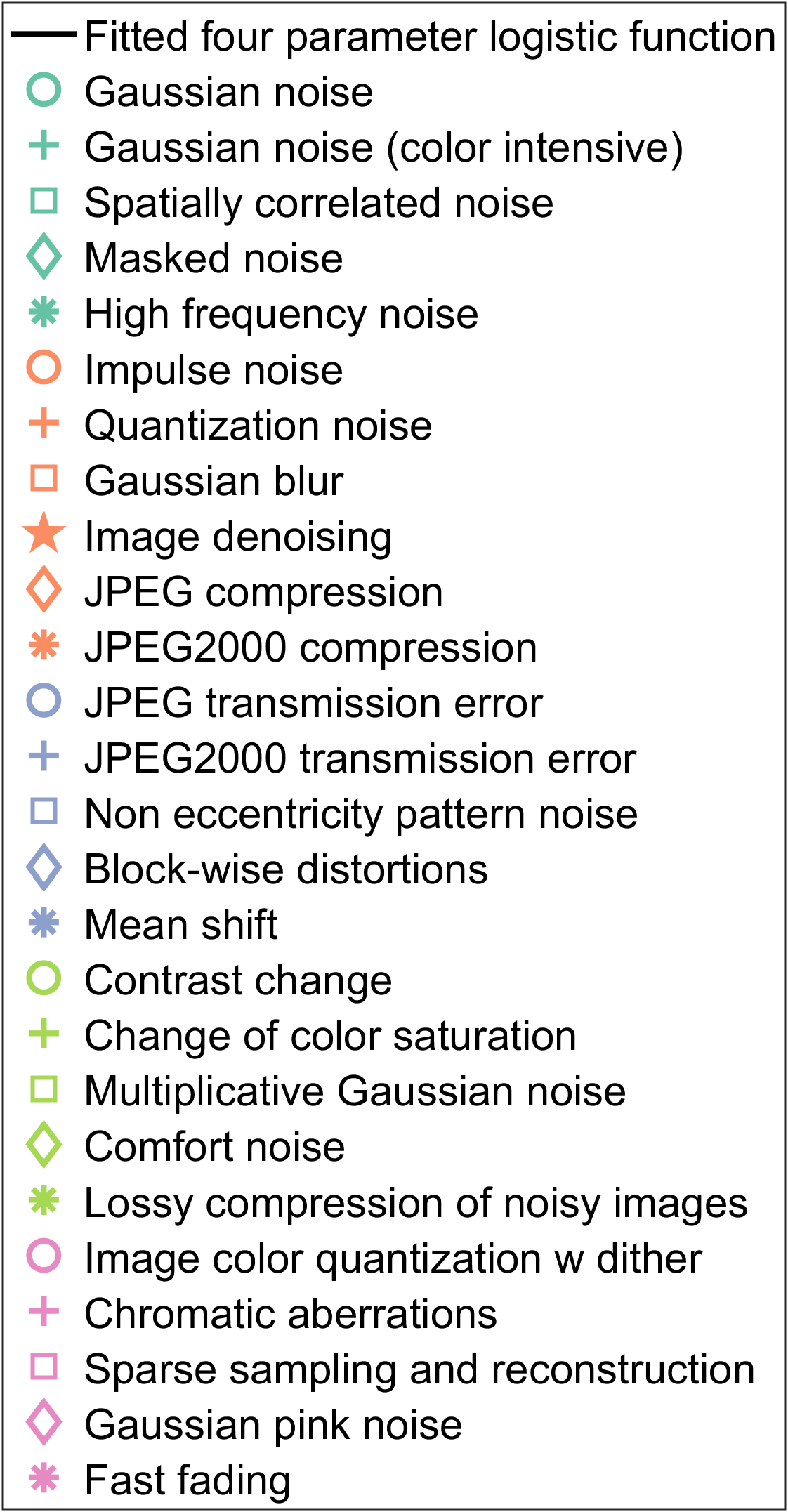}
	\end{minipage}\hfil
	\begin{minipage}{0.6\linewidth}
		\subfloat[LIVE]{\includegraphics[width = 0.45\textwidth]{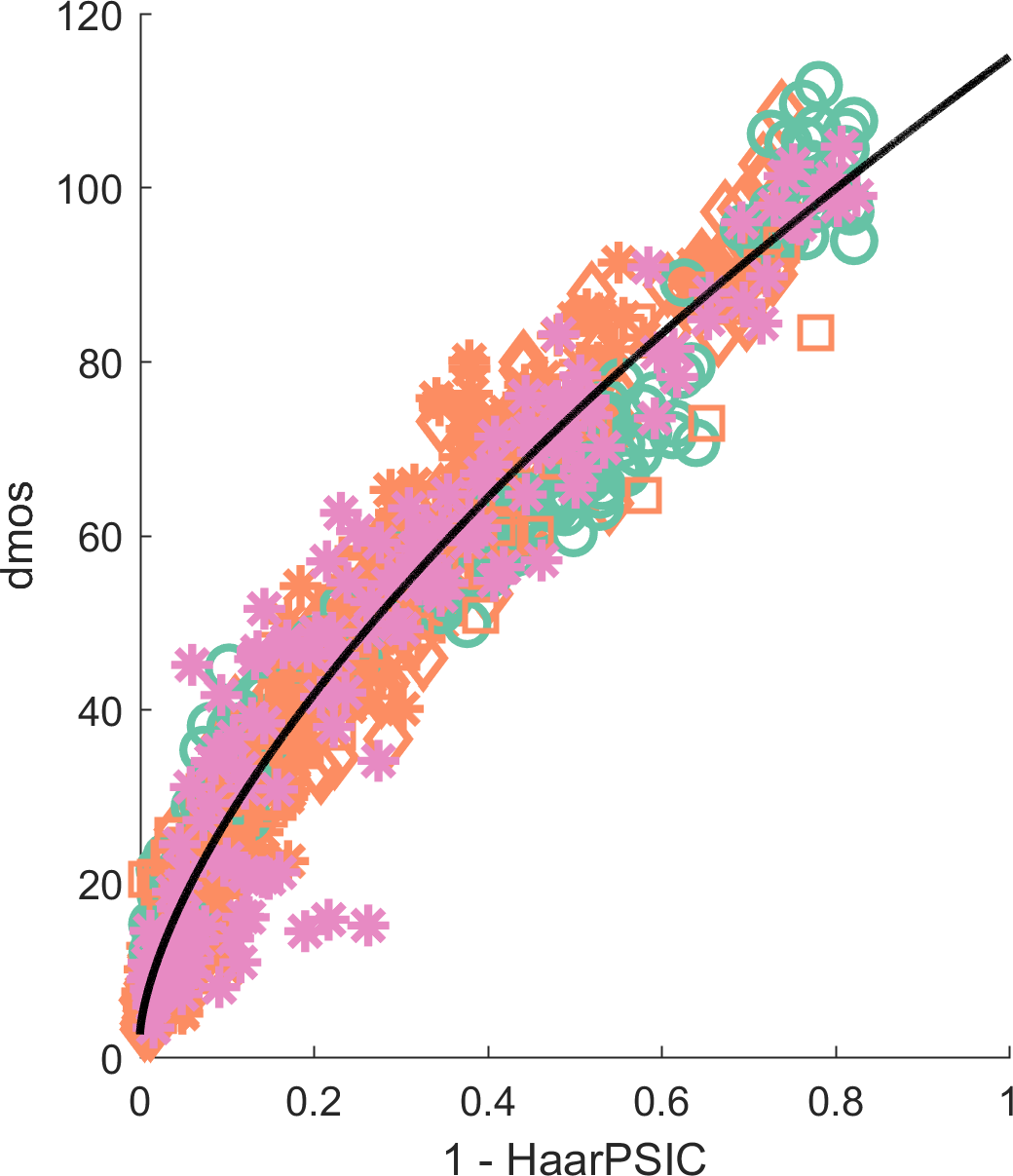}}\hfil\subfloat[TID 2008]{\includegraphics[width = 0.45\textwidth]{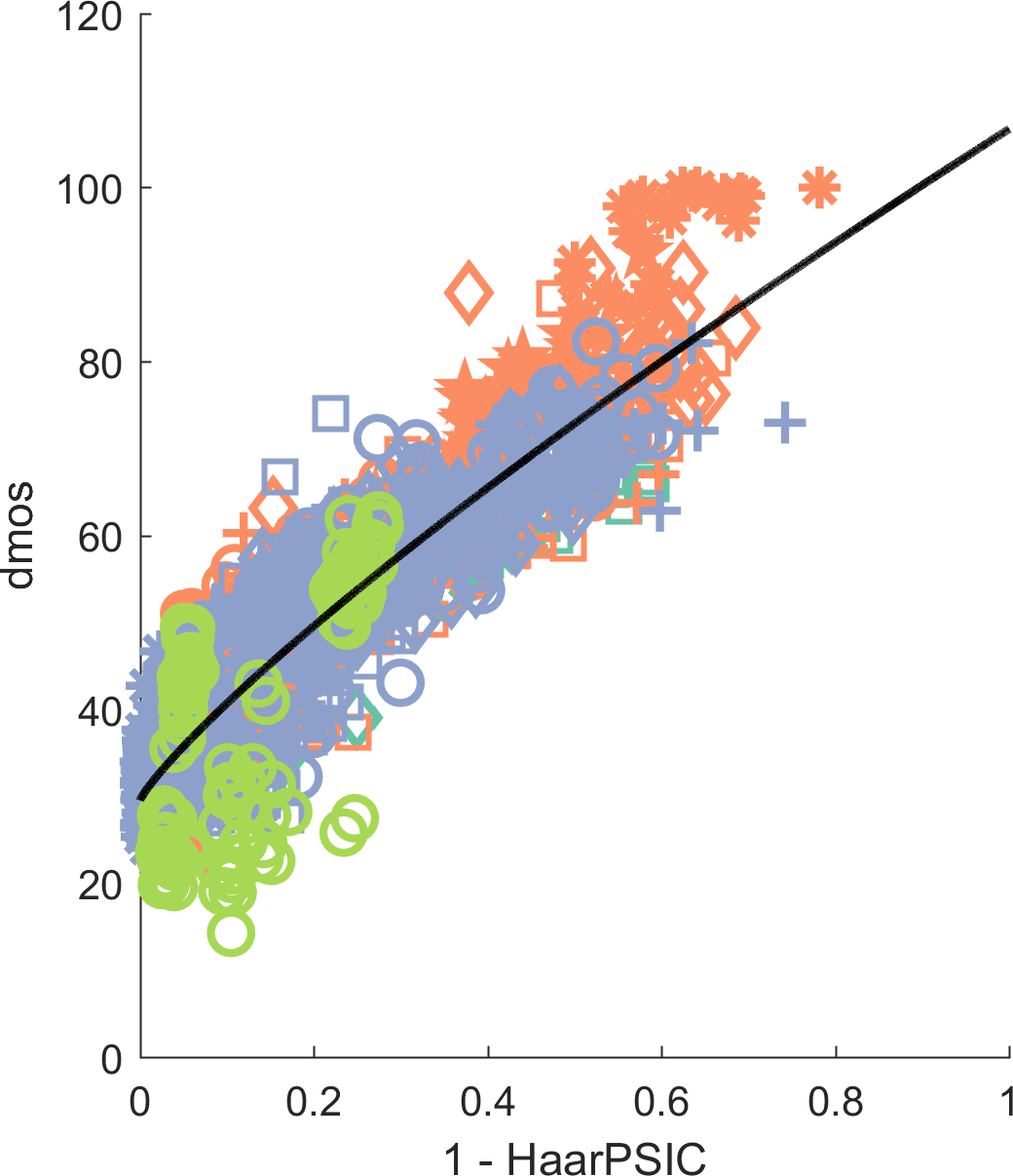}}\\\subfloat[TID 2013]{\includegraphics[width = 0.45\textwidth]{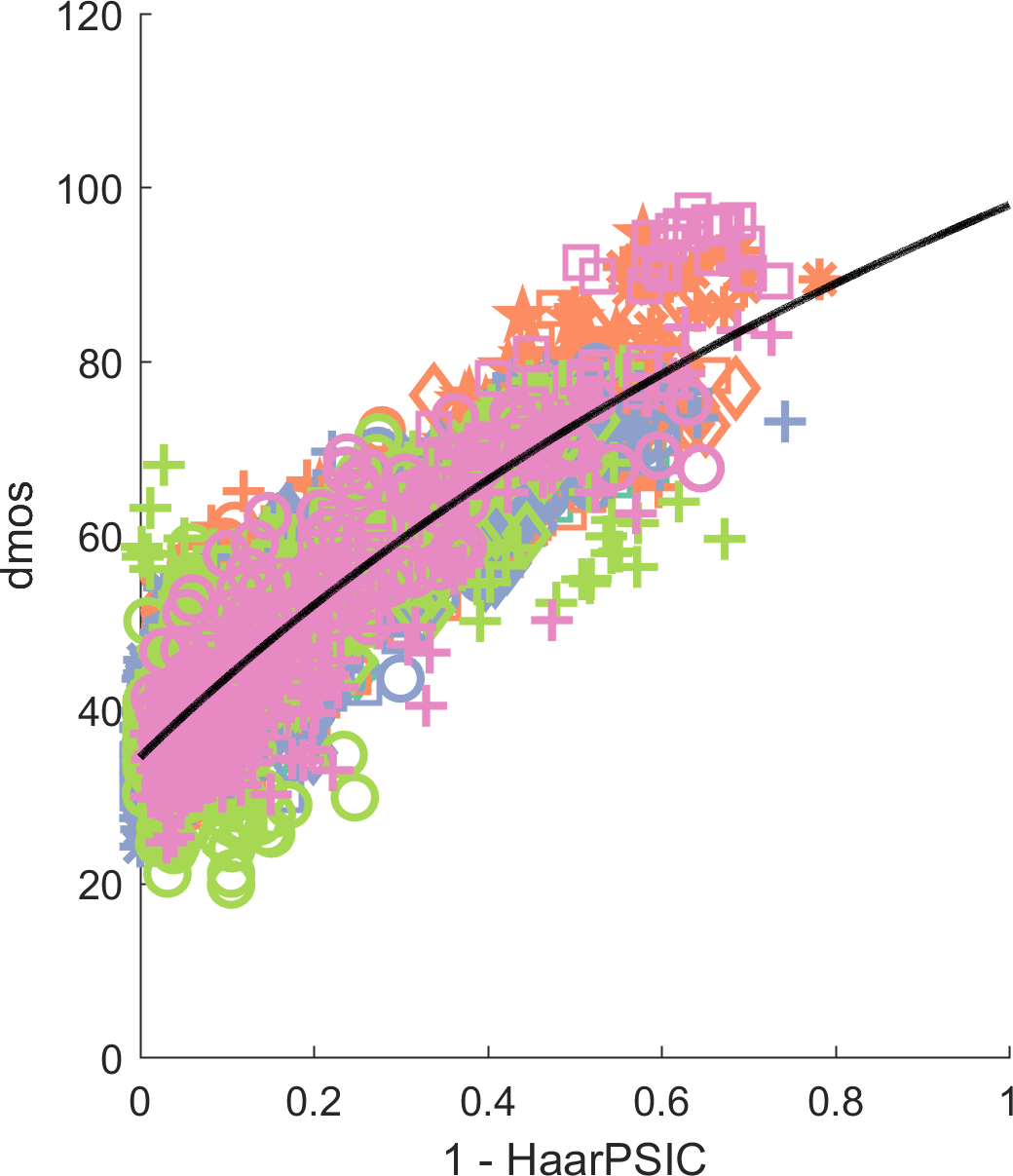}}\hfil\subfloat[CSIQ]{\includegraphics[width = 0.45\textwidth]{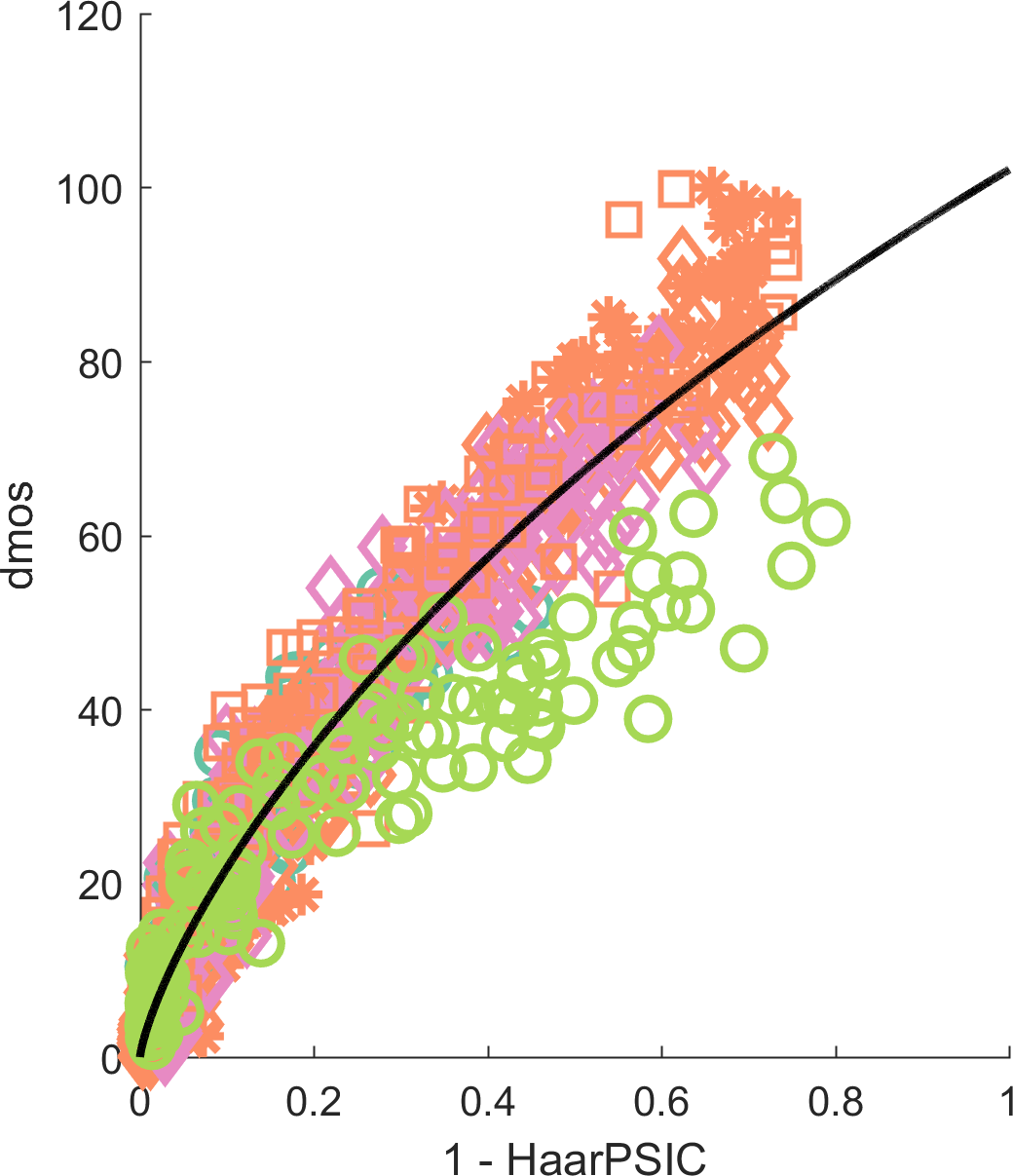}}
	\end{minipage}
	\caption{Scatter plots of HaarPSIC values against difference mean opinions scores (DMOS)  from the LIVE, TID 2008, TID 2013 and CSIQ image databases.}
	\label{fig:scatterplots}
\end{figure}

%text about pearson tables an liear relationship (combined with scatterplots)
Single correlation coefficients provide a useful means of objectively evaluating and comparing different computational models of image quality. However, they only measure a specific aspect of the relationship between an image similarity metric and human opinion scores, like linearity in the case of the Pearson correlation coefficient or monotonicity in the case of the SROCC. In an attempt to better visualize the relationship between the HaarPSI and human opinion scores, Figure~\ref{fig:scatterplots} shows scatter plots of the HaarPSI against difference mean opinion scores (DMOS) for all four databases. To provide as much insight as possible, the plots are categorized by specific distortion types.

\FloatBarrier
\setlength{\tabcolsep}{1mm}
\begin{table}[!htb]
	\centering
	\caption{Spearman Rank Order Correlations of IQA Metrics With Human Mean Opinion Scores}
	\label{tab:details}
	\begin{scriptsize}
		\begin{threeparttable}
			\begin{tabular}{*{12}{c}}
				\toprule[0.5mm]
				\multicolumn{12}{c}{Color Images}\\[0.1cm]
				& & PSNR & VIF & SSIM & MS-SSIM & GSM & MAD & SR-SIM & FSIM & VSI & HaarPSI\\
				\multirow{5}{*}{LIVE} & jpg2k & \cellcolor{green!25}0.8954 & 0.9696 & 0.9614 & 0.9627 & 0.9700 & 0.9692 & 0.9700 & \textbf{0.9724} & 0.9604 & 0.9684\\
				& jpg & \cellcolor{green!25}0.8809 & \textbf{0.9846} & 0.9764 & 0.9815 & 0.9778 & 0.9786 & 0.9823 & 0.9840 & 0.9761 & 0.9832\\
				& gwn & 0.9854 & 0.9858 & \cellcolor{green!25}0.9694 & \cellcolor{green!25}0.9733 & 0.9774 & \textbf{0.9873} & 0.9812 & \cellcolor{green!25}0.9716 & 0.9835 & 0.9845\\
				& gblur & \cellcolor{green!25}0.7823 & \textbf{0.9728} & 0.9517 & 0.9542 & 0.9518 & 0.9510 & 0.9660 & 0.9708 & 0.9527 & 0.9676\\
				& ff & \cellcolor{green!25}0.8907 & \textbf{0.9650} & 0.9556 & 0.9471 & 0.9402 & 0.9589 & 0.9466 & 0.9519 & 0.9430 & 0.9527\\
				\\
				\multirow{17}{*}{TID2008} & gwn & 0.9070 & 0.8797 & \cellcolor{green!25}0.8107 & \cellcolor{green!25}0.8086 & 0.8606 & \cellcolor{green!25}0.8386 & 0.8989 & 0.8758 & \textbf{0.9229} & 0.9177\\
				& gwnc & 0.8995 & 0.8757 & \cellcolor{green!25}0.8029 & \cellcolor{green!25}0.8054 & \cellcolor{green!25}0.8091 & 0.8255 & 0.8957 & 0.8931 & \textbf{0.9118} & 0.8982\\
				& scn & 0.9170 & \cellcolor{green!25}0.8698 & \cellcolor{green!25}0.8145 & \cellcolor{green!25}0.8209 & 0.8941 & \cellcolor{green!25}0.8678 & 0.9084 & \cellcolor{green!25}0.8711 & \textbf{0.9296} & 0.9271\\
				& mn & 0.8515 & \textbf{0.8683} & 0.7795 & 0.8107 & 0.7452 & 0.7336 & 0.7881 & 0.8264 & 0.7734 & 0.7909\\
				& hfn & \textbf{0.9270} & 0.9075 & 0.8729 & 0.8694 & 0.8945 & 0.8864 & 0.9195 & 0.9156 & 0.9253 & 0.9155\\
				& in & \textbf{0.8724} & 0.8327 & \cellcolor{green!25}0.6732 & \cellcolor{green!25}0.6907 & 0.7235 & \cellcolor{green!25}0.0650 & 0.7678 & 0.7719 & 0.8298 & 0.8269\\
				& qn & 0.8696 & \cellcolor{green!25}0.7970 & 0.8531 & 0.8589 & 0.8800 & 0.8160 & 0.8348 & 0.8726 & 0.8731 & \textbf{0.8842}\\
				& gblr & 0.8697 & \cellcolor{red!25}0.9540 & \cellcolor{red!25}0.9544 & \cellcolor{red!25}0.9563 & \cellcolor{red!25}\textbf{0.9600} & 0.9196 & \cellcolor{red!25}0.9551 & \cellcolor{red!25}0.9472 & \cellcolor{red!25}0.9529 & 0.9001\\
				& den & \cellcolor{green!25}0.9416 & \cellcolor{green!25}0.9161 & 0.9530 & 0.9582 & \textbf{0.9725} & \cellcolor{green!25}0.9433 & 0.9666 & 0.9618 & 0.9693 & 0.9711\\
				& jpg & \cellcolor{green!25}0.8717 & 0.9168 & 0.9252 & 0.9322 & 0.9393 & 0.9275 & 0.9393 & 0.9294 & \textbf{0.9616} & 0.9417\\
				& jpg2k & \cellcolor{green!25}0.8132 & \cellcolor{green!25}0.9709 & \cellcolor{green!25}0.9625 & \cellcolor{green!25}0.9700 & 0.9758 & \cellcolor{green!25}0.9707 & 0.9809 & 0.9780 & 0.9848 & \textbf{0.9860}\\
				& jpgt & \cellcolor{green!25}0.7516 & 0.8585 & 0.8678 & 0.8681 & 0.8790 & 0.8661 & 0.8881 & 0.8756 & \textbf{0.9160} & 0.8921\\
				& jpg2kt & 0.8309 & 0.8501 & 0.8577 & 0.8606 & 0.8936 & 0.8394 & 0.8902 & 0.8555 & 0.8942 & \textbf{0.8963}\\
				& pn & \cellcolor{green!25}0.5815 & 0.7619 & 0.7107 & 0.7377 & 0.7386 & \textbf{0.8287} & 0.7659 & 0.7514 & 0.7699 & 0.8010\\
				& bdist & \cellcolor{green!25}0.6193 & 0.8324 & 0.8462 & 0.7546 & \cellcolor{red!25}\textbf{0.8862} & 0.7970 & 0.7798 & 0.8464 & \cellcolor{green!25}0.6295 & 0.8026\\
				& ms & 0.6957 & 0.5096 & 0.7231 & \textbf{0.7338} & 0.7190 & 0.5163 & 0.5704 & 0.6554 & 0.6714 & 0.6051\\
				& ctrst & 0.5859 & \cellcolor{red!25}\textbf{0.8188} & 0.5246 & 0.6381 & 0.6691 & \cellcolor{green!25}0.2723 & 0.6475 & 0.6510 & 0.6557 & 0.6209\\
				\\
				\multirow{24}{*}{TID2013} & gwn & 0.9291 & 0.8994 & \cellcolor{green!25}0.8671 & \cellcolor{green!25}0.8646 & 0.9064 & \cellcolor{green!25}0.8843 & 0.9251 & 0.9101 & \textbf{0.9460} & 0.9368\\
				& gwnc & \textbf{0.8981} & 0.8299 & \cellcolor{green!25}0.7726 & \cellcolor{green!25}0.7730 & 0.8175 & 0.8019 & 0.8562 & 0.8537 & 0.8705 & 0.8593\\
				& scn & 0.9200 & \cellcolor{green!25}0.8835 & \cellcolor{green!25}0.8515 & \cellcolor{green!25}0.8544 & 0.9158 & 0.8911 & 0.9223 & 0.8900 & \textbf{0.9367} & 0.9311\\
				& mn & 0.8323 & \textbf{0.8450} & 0.7767 & 0.8073 & 0.7293 & 0.7380 & 0.7855 & 0.8094 & 0.7697 & 0.7858\\
				& hfn & 0.9140 & 0.8972 & 0.8634 & 0.8604 & 0.8869 & 0.8876 & 0.9131 & 0.9040 & \textbf{0.9200} & 0.9069\\
				& in & \textbf{0.8968} & 0.8537 & \cellcolor{green!25}0.7503 & \cellcolor{green!25}0.7629 & 0.7965 & \cellcolor{green!25}0.2769 & 0.8280 & 0.8251 & 0.8741 & 0.8656\\
				& qn & 0.8808 & \cellcolor{green!25}0.7854 & 0.8657 & 0.8706 & 0.8841 & 0.8514 & 0.8497 & 0.8807 & 0.8748 & \textbf{0.8893}\\
				& gblr & 0.9149 & \cellcolor{red!25}0.9650 & \cellcolor{red!25}0.9668 & \cellcolor{red!25}0.9673 & \cellcolor{red!25}\textbf{0.9689} & 0.9319 & \cellcolor{red!25}0.9622 & \cellcolor{red!25}0.9551 & \cellcolor{red!25}0.9612 & 0.9149\\
				& den & 0.9480 & \cellcolor{green!25}0.8911 & 0.9254 & 0.9268 & 0.9432 & 0.9252 & 0.9398 & 0.9330 & \textbf{0.9484} & 0.9456\\
				& jpg & \cellcolor{green!25}0.9189 & \cellcolor{green!25}0.9192 & 0.9200 & 0.9265 & 0.9284 & 0.9217 & 0.9396 & 0.9339 & \textbf{0.9541} & 0.9512\\
				& jpg2k & \cellcolor{green!25}0.8840 & 0.9516 & \cellcolor{green!25}0.9468 & \cellcolor{green!25}0.9504 & 0.9602 & 0.9511 & 0.9672 & 0.9589 & \textbf{0.9706} & 0.9704\\
				& jpgt & \cellcolor{green!25}0.7685 & 0.8409 & 0.8493 & 0.8475 & 0.8512 & 0.8283 & 0.8543 & 0.8610 & \textbf{0.9216} & 0.8938\\
				& jpg2kt & 0.8883 & 0.8761 & 0.8828 & 0.8889 & 0.9182 & 0.8788 & 0.9165 & 0.8919 & \textbf{0.9228} & 0.9204\\
				& pn & \cellcolor{green!25}0.6863 & 0.7720 & 0.7821 & 0.7968 & 0.8130 & \textbf{0.8315} & 0.7967 & 0.7937 & 0.8060 & 0.8154\\
				& bdist & \cellcolor{green!25}0.1552 & 0.5306 & 0.5720 & 0.4801 & \cellcolor{red!25}\textbf{0.6418} & 0.2812 & 0.4722 & 0.5532 & \cellcolor{green!25}0.1713 & 0.4471\\
				& ms & 0.7671 & 0.6276 & 0.7752 & \textbf{0.7906} & 0.7875 & 0.6450 & 0.6562 & 0.7487 & 0.7700 & 0.7152\\
				& ctrst & 0.4400 & \cellcolor{red!25}\textbf{0.8386} & 0.3775 & 0.4634 & 0.4857 & \cellcolor{green!25}0.1972 & 0.4696 & 0.4679 & 0.4754 & 0.4382\\
				& ccs & \cellcolor{green!25}0.0766 & \cellcolor{green!25}0.3099 & \cellcolor{green!25}0.4141 & \cellcolor{green!25}0.4099 & \cellcolor{green!25}0.3578 & \cellcolor{green!25}0.0575 & \cellcolor{green!25}0.3117 & \cellcolor{red!25}\textbf{0.8359} & \cellcolor{red!25}0.8100 & 0.6735\\
				& mgn & 0.8905 & 0.8468 & \cellcolor{green!25}0.7803 & \cellcolor{green!25}0.7786 & 0.8348 & 0.8409 & 0.8781 & 0.8569 & \textbf{0.9117} & 0.8902\\
				& cn & \cellcolor{green!25}0.8411 & 0.8946 & \cellcolor{green!25}0.8566 & \cellcolor{green!25}0.8528 & 0.9124 & 0.9064 & 0.9259 & 0.9135 & 0.9243 & \textbf{0.9275}\\
				& lcni & \cellcolor{green!25}0.9145 & \cellcolor{green!25}0.9204 & \cellcolor{green!25}0.9057 & \cellcolor{green!25}0.9068 & 0.9563 & 0.9443 & 0.9608 & 0.9485 & 0.9564 & \textbf{0.9622}\\
				& icqd & \textbf{0.9269} & 0.8414 & 0.8542 & 0.8555 & 0.8973 & 0.8745 & 0.8810 & 0.8815 & 0.8839 & 0.8953\\
				& cha & 0.8872 & 0.8848 & 0.8775 & 0.8784 & 0.8823 & 0.8310 & 0.8758 & \textbf{0.8925} & 0.8906 & 0.8599\\
				& ssr & \cellcolor{green!25}0.9042 & \cellcolor{green!25}0.9353 & 0.9461 & 0.9483 & \textbf{0.9668} & 0.9567 & 0.9613 & 0.9576 & 0.9628 & 0.9651\\
				\\
				\multirow{6}{*}{CSIQ} & gwn & \cellcolor{green!25}0.9363 & 0.9575 & \cellcolor{green!25}0.8974 & 0.9471 & \cellcolor{green!25}0.9440 & 0.9541 & 0.9628 & \cellcolor{green!25}0.9359 & 0.9636 & \textbf{0.9666}\\
				& jpeg & \cellcolor{green!25}0.8881 & \textbf{0.9705} & 0.9546 & 0.9634 & 0.9632 & 0.9615 & 0.9671 & 0.9664 & 0.9618 & 0.9695\\
				& jpg2k & \cellcolor{green!25}0.9362 & \cellcolor{green!25}0.9672 & \cellcolor{green!25}0.9606 & \cellcolor{green!25}0.9683 & \cellcolor{green!25}0.9648 & 0.9752 & 0.9773 & \cellcolor{green!25}0.9704 & \cellcolor{green!25}0.9694 & \textbf{0.9815}\\
				& gpn & \cellcolor{green!25}0.9339 & 0.9511 & \cellcolor{green!25}0.8922 & \cellcolor{green!25}0.9331 & 0.9387 & 0.9570 & 0.9520 & 0.9370 & \textbf{0.9638} & 0.9594\\
				& gblr & \cellcolor{green!25}0.9291 & 0.9745 & \cellcolor{green!25}0.9609 & 0.9711 & \cellcolor{green!25}0.9589 & 0.9682 & 0.9767 & 0.9729 & 0.9679 & \textbf{0.9783}\\
				& ctrst & \cellcolor{green!25}0.8621 & 0.9345 & \cellcolor{green!25}0.7922 & 0.9526 & 0.9354 & 0.9207 & \textbf{0.9528} & 0.9438 & 0.9504 & 0.9450\\\midrule[0.5mm]
			\end{tabular}
			\begin{tablenotes}
				\item \colorbox{green!25}{Lower correlation than HaarPSI. The difference is statistically significant with $p < 0.05$.}
				\item \colorbox{red!25}{Higher correlation than HaarPSI. The difference is statistically significant with $p < 0.05$.}
				\item The highest correlation in each row is written in \textbf{boldface}.
			\end{tablenotes}
		\end{threeparttable}
	\end{scriptsize}
\end{table}

%Preprocessing with subsampling
It should be noted that for all results reported in this section,
the HaarPSI, as well as other image quality metrics such as the SSIM, the FSIM or the VSI, were preprocessing each image by convolving it with a $2\times2$ mean filter as well as a subsequent dyadic subsampling step. This preprocessing approximates the
low-pass characteristics of the optical part of the human visual system
\cite{palmer1999vision} by a simple model.

\section{Conclusion}

%basics
The HaarPSI is a novel and computationally inexpensive image quality measure based solely on the coefficients of three stages of a discrete Haar wavelet transform. Its validity with respect to the human perception of image quality was tested on four large databases containing more than 5000 differently distorted images, with very promising results. In a comparison with 9 popular state-of-the-art image similarity metrics, the HaarPSI yields significantly higher or statistically indistinguishable Spearman correlations when restricted to grayscale conversions. For color images, it only comes second to the VSI when tested on the TID 2013 (see Table~\ref{tab:sroccdatabases}). Along with its simple computational structure and its comparatively short execution time, this suggests a high applicability of the HaarPSI in real world optimization tasks. In particular, image quality metrics like PSNR, SSIM, or SR-SIM, that outperform the HaarPSI with respect to speed achieve considerably inferior correlations with human opinion scores (see Table~\ref{tab:overallperf}). Regarding the applicability of the HaarPSI in specific optimization tasks, we would like to mention that the HaarPSI has consistently high correlations with human opinion scores throughout all databases with respect to distortions caused by the JPEG and JPEG 2000 compression algorithms (see Table~\ref{tab:details}).  

%highest overall correlation vs. being outperformed in
The results reported in Tables~\ref{tab:sroccdatabases}~and~\ref{tab:details} might seem contradictory at first glance. In many cases, the HaarPSI yields the highest SROCC for a complete database but is outperformed by other metrics like the VSI when restricting the same database to a single distortion type. However, taking into account statistical significance, it is apparent that only when tested on the TID databases restricted to Gaussian blur, the performance of the HaarPSI is consistently lower than the performance of other similarity metrics. This particular shortcoming can be explained by the fact that the HaarPSI is almost exclusively relying on high-frequency information and thus maybe too sensitive in the case of distortions purely based on low-pass filtering.

\begin{figure}[!htb]
	\centering
	\subfloat[JPEG]{\label{fig:corSpearVsAlphaC1_jpeg}\includegraphics[width=0.25\textwidth]{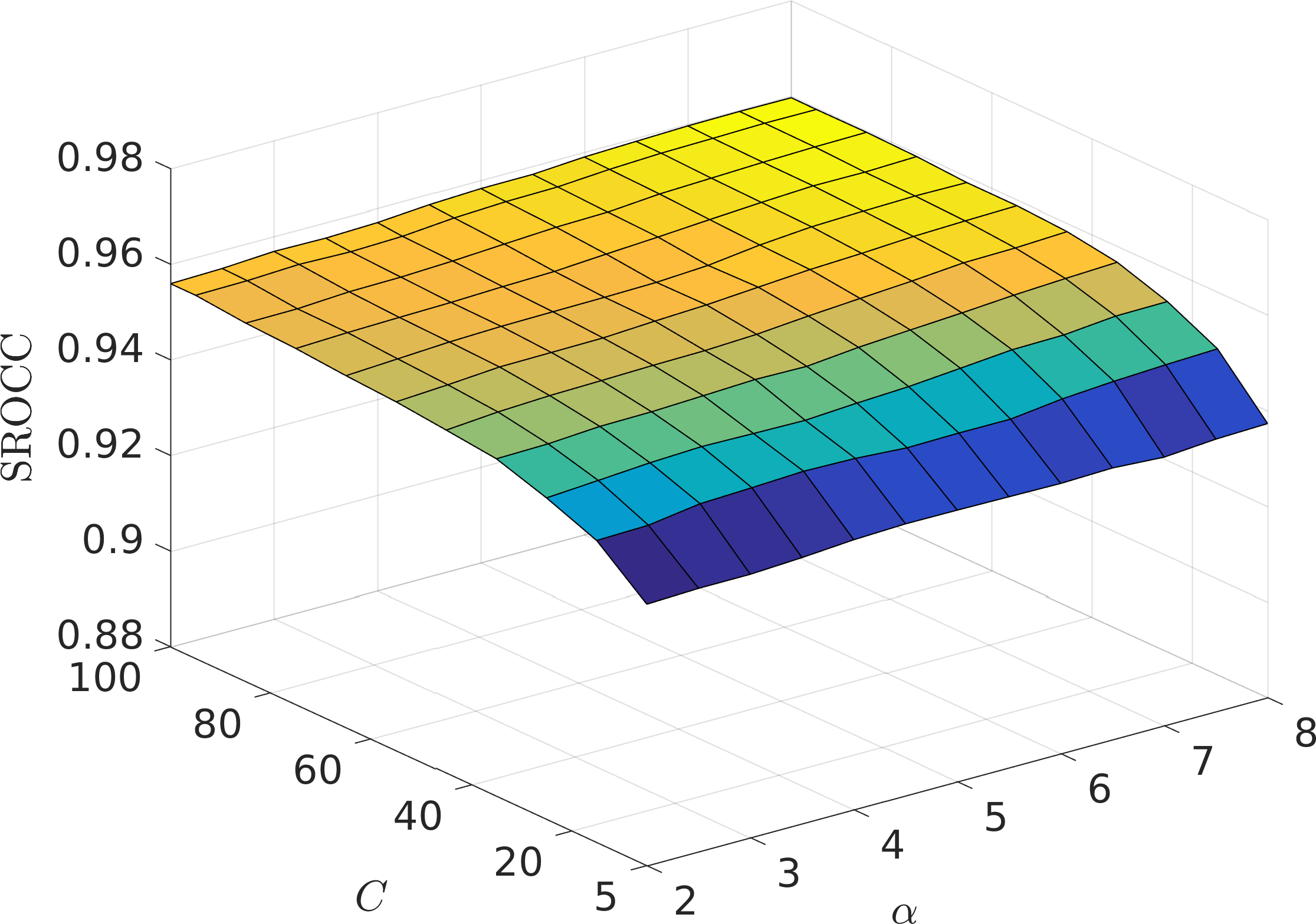}}\hfil
	\subfloat[JP2K]{\label{fig:corSpearVsAlphaC1_jp2k}\includegraphics[width=0.25\textwidth]{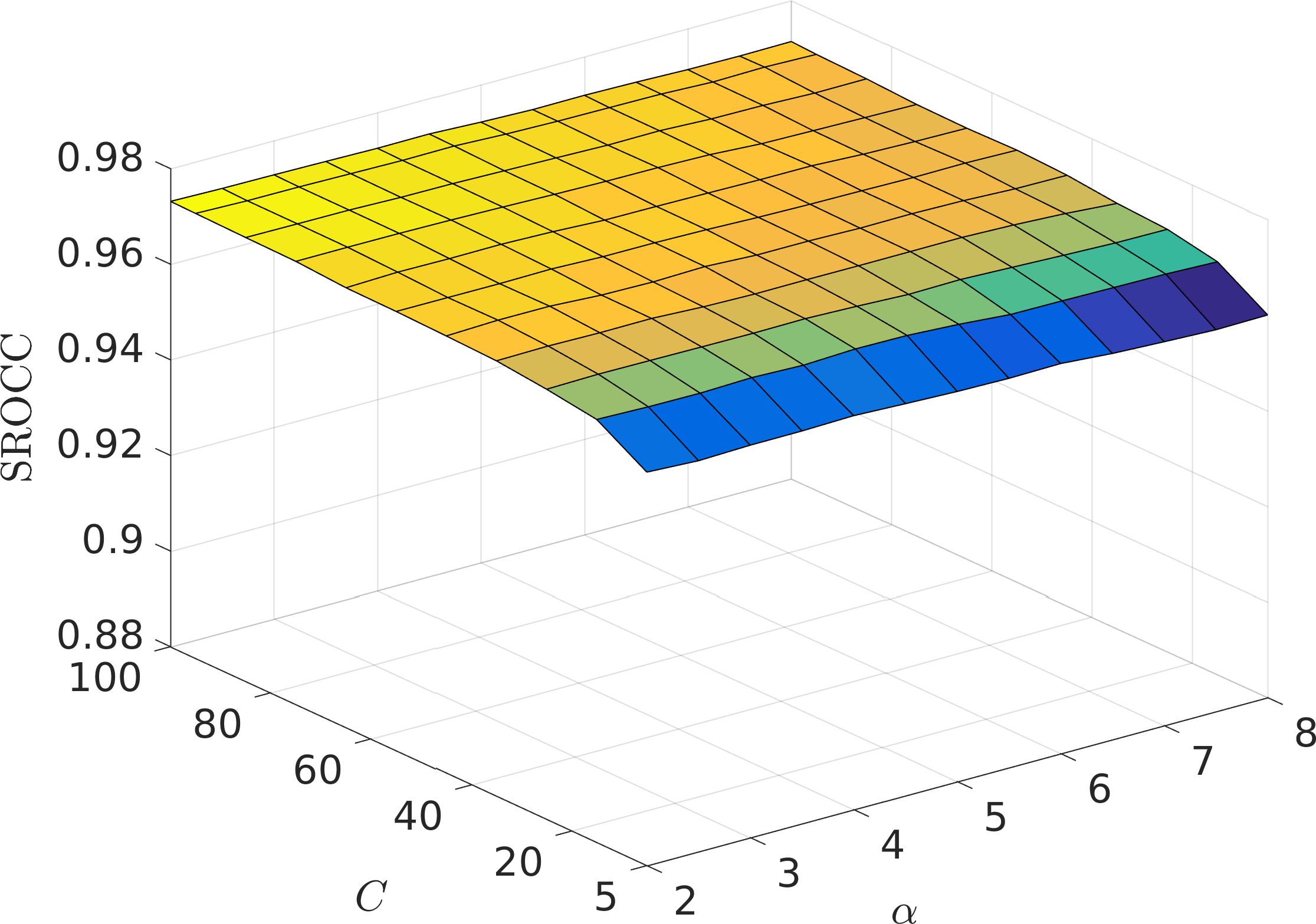}}\hfil
	\subfloat[Blur]{\label{fig:corSpearVsAlphaC1_blur}\includegraphics[width=0.25\textwidth]{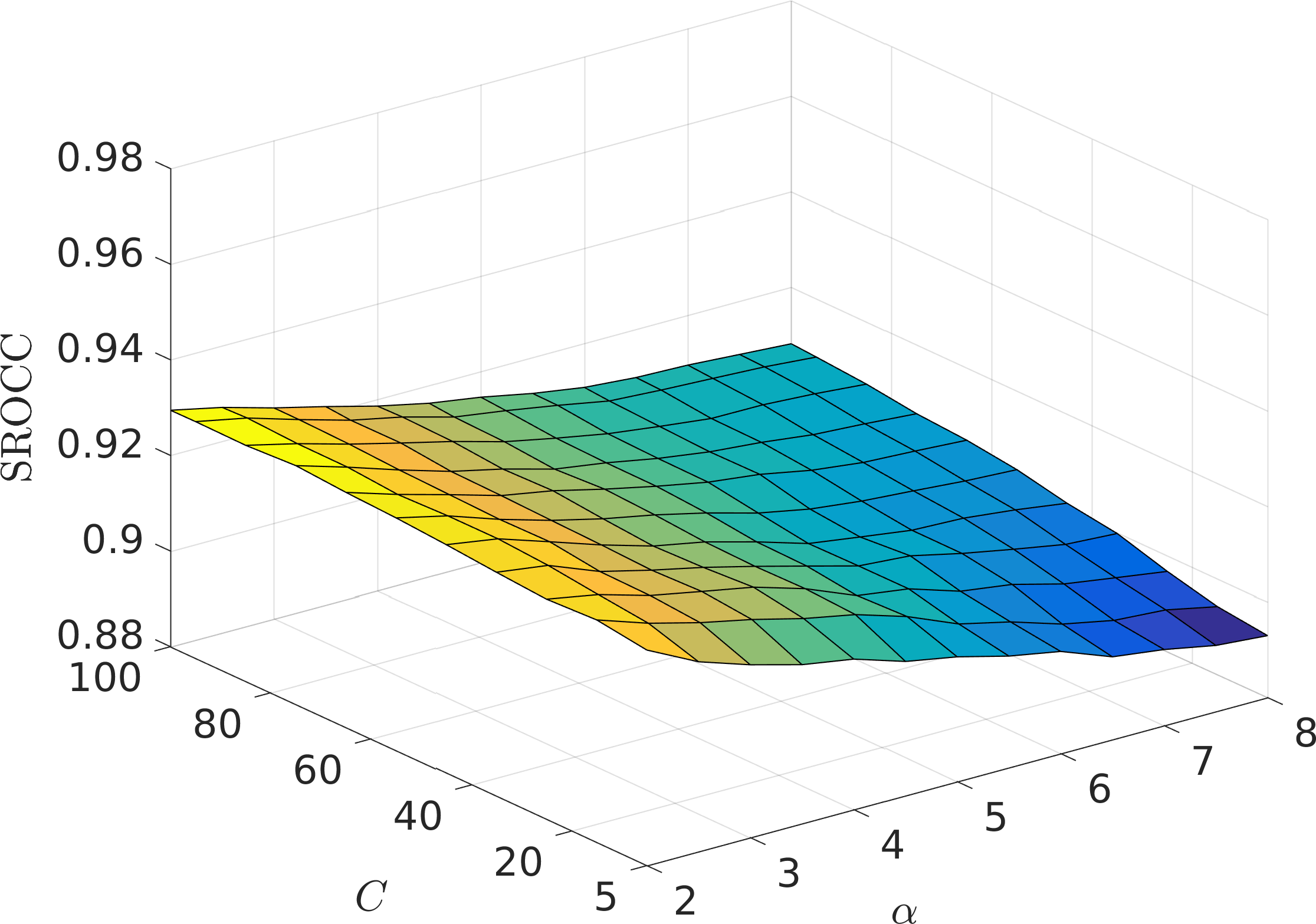}}\\
	\subfloat[AWGN]{\label{fig:corSpearVsAlphaC1_awgn}\includegraphics[width=0.25\textwidth]{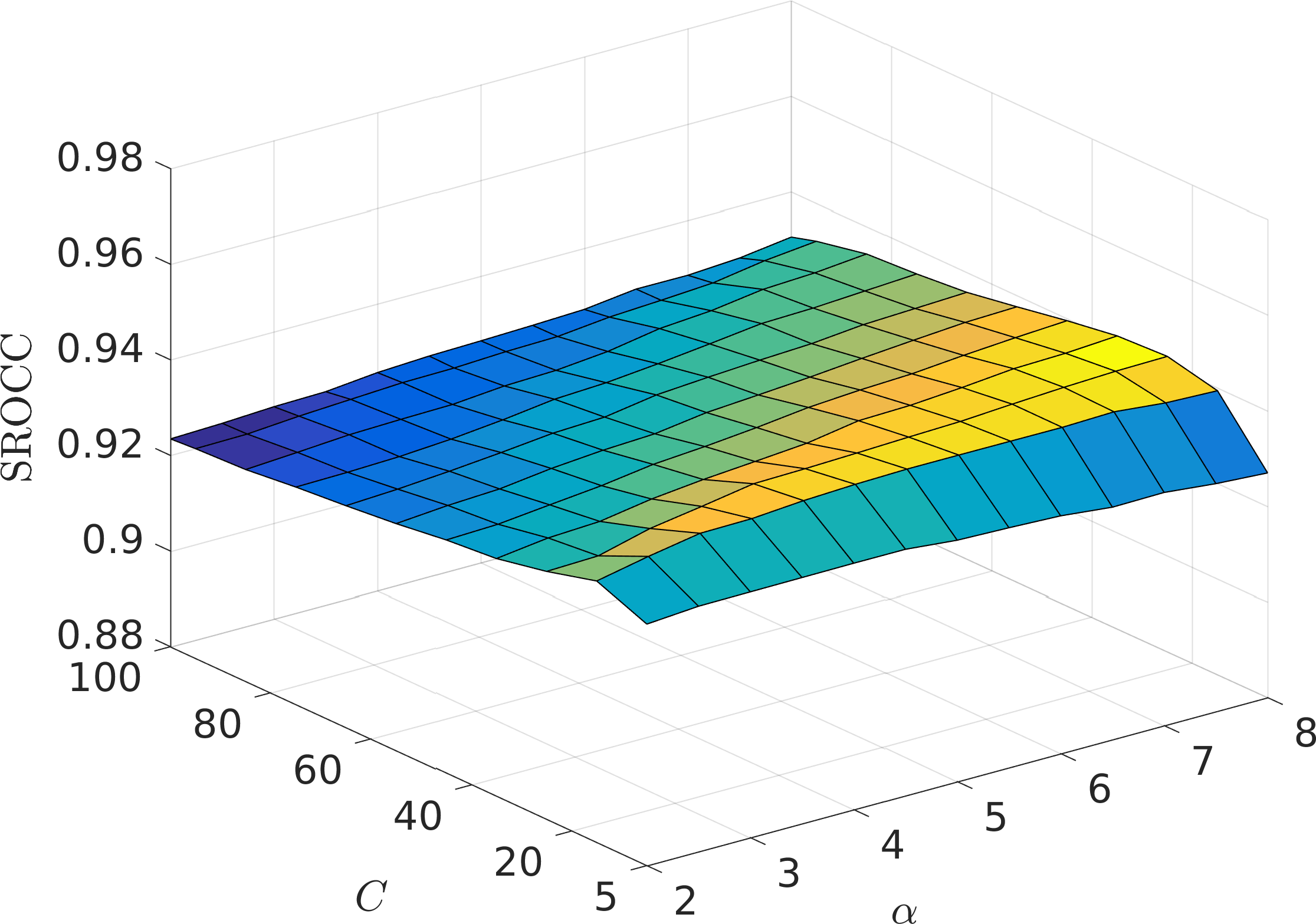}}\hfil
	\subfloat[SCN]{\label{fig:corSpearVsAlphaC1_scn}\includegraphics[width=0.25\textwidth]{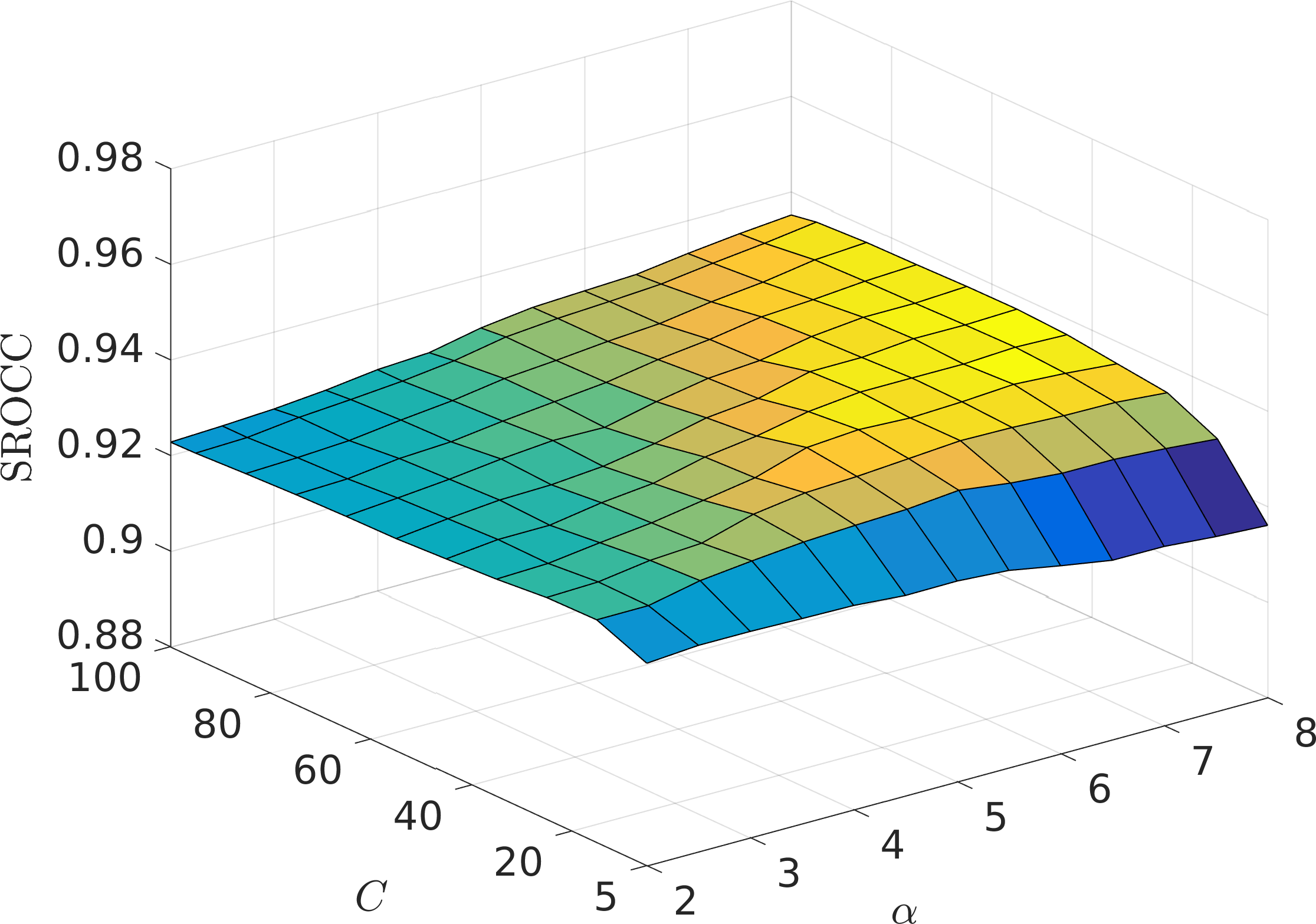}}\hfil
	\subfloat[Denoising]{\label{fig:corSpearVsAlphaC1_all}\includegraphics[width=0.25\textwidth]{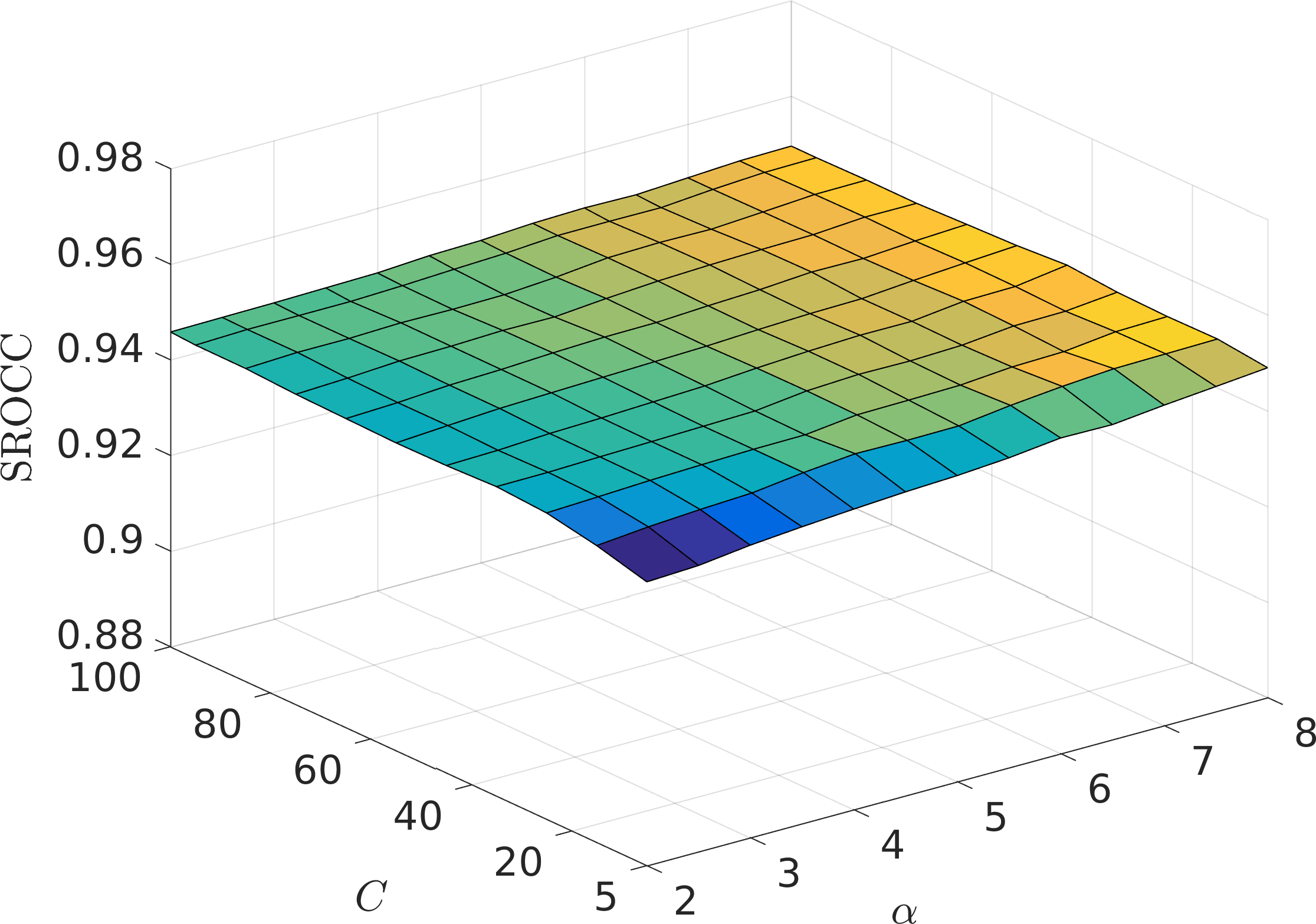}}\hfil
	%   \caption{Spearman correlation $r_S$ in dependance of parameters $C1$ and
	%   $\alpha$ for images affecty by (a) all distortions, (b) JPEG compression, (c)
	%   JP2K compression, (d) Gaussian Blur, (e) additive Gaussien white
	%   noise, (f) high frequency noise, (g) spatially correlated noise.
	%   All correlations are obtained based on TID2013.}
	\caption{Spearman rank order correlations as functions of the parameters $C$ and
		$\alpha$ for images affected by (a) JPEG compression, (b) JP2K compression, (c)
			Gaussian Blur, (d) additive Gaussien white
			noise, (e) spatially correlated noise white
			noise, and (f) denoising. All correlations are with respect to TID2013.}
	\label{fig:corSpearVsAlphaC1}
\end{figure}

%parameter tuning for specific distortions
When only considering a specific type of distortion, the correlations yielded by the HaarPSI might be improved by tuning the constants $C$ and $\alpha$, which have originally been selected to optimize the overall performance. Increasing $C$ decreases the sensitivity of the HaarPSI to changes in the high-frequency components measured by the similarity maps $\operatorname{HS^\text{(1,2)}_{f_1,f_2}}$ relative to the weights $\operatorname{W^\text{(1,2)}_{f}}$, which are based on a lower frequency band and serve as a rough model of attention-like processes. The effect of the parameter $\alpha$ on the HaarPSI is qualitatively similar when it is approaching zero. This could explain the roughly negative linear relationship between $C$ and $\alpha$ in Figure~\ref{fig:optimization}. However, for larger choices of $\alpha$, the function $l_\alpha(\cdot)$ is increasingly mimicking the behavior of a thresholding operator in the sense that only severe changes in the high-frequency components will have a significant effect on the HaarPSI. To also provide a quantitative analysis of these relationships, Figure~\ref{fig:corSpearVsAlphaC1} depicts the influence of $C$ and $\alpha$ on the correlation with human opinion scores in the case of TID 2013 with respect to six different distortions. Figure~\ref{fig:corSpearVsAlphaC1_blur} indeed suggests that in the case of Gaussian blur, the performance of the HaarPSI can be improved by attenuating its sensitivity to changes in the high-frequency components via increasing $C$ and choosing $\alpha$ close to 0. In contrast, Figure~\ref{fig:corSpearVsAlphaC1_jpeg} indicates that the HaarPSI achieves the highest correlations in the case of JPEG compression artifacts when it is tuned to be sensitive to severe changes in the high frequency components at highly salient locations.

%neurophysiologically not so meaningful
{It is surprising that the extremely simple computational model of orientation and spatial frequency selectivity used in the HaarPSI suffices to obtain comparatively high correlations with human opinion scores. Additionally, these correlations are stable with respect to a wide range of parameters $C$ and $\alpha$ (cf. Figure~\ref{fig:optimization}). This could indicate that the computational structure of the HaarPSI succeeds at reproducing the \textit{functional essence} of at least some parts of the human visual system. It is, however, quite likely that the HaarPSI owes some of its experimental success to the limitations of the used benchmark databases, which only consider a limited number of reference images and specific types of distortions. Certainly, orientation selectivity in the primary visual cortex is not restricted to horizontal and vertical edges.}

{Another computational principle that plays an important role in natural neural systems and that was recently successfully applied in the context of perceptual image similarity measurement is \textit{divisive normalization} \cite{laparra2017perceptually}. While the similarity measure $\Sim(a,b,C)$ introduces some kind of normalization, divisive normalization is not included in any of the computational stages of the HaarPSI. It remains an open question if and how the HaarPSI could be further improved by incorporating divisive normalization in a similar fashion as the concepts of orientation selectivity and spatial frequency selectivity.}

%interpretability/lineartiy
Many practical applications demand image similarity metrics to yield values that are easy to interpret. Ideally, an image similarity of $0.9$ would in fact indicate that the average human would also assess a similarity of $\SI{90}{\%}$ between two images or that a decrease in similarity to $0.8$ corresponds to a $\SI{10}{\%}$ decrease in perceived quality for a human viewer. Due to the generality and difficulty of this task, computational models of image similarity typically only aim at establishing a monotonic relationship with human mean opinion scores, which is also reflected in the choice of the SROCC as a measure of consistency. In the case of the HaarPSI, applying $l_\alpha^{-1}(\cdot)^2$ to the final similarity score significantly linearizes its relationship with human opinion scores, thereby leading to the strong linear correlations depicted in the scatter plots in Figure~\ref{fig:scatterplots}. While $l_\alpha^{-1}(\cdot)^2$ is monotonically increasing on $[\frac{1}{2},1)$ and therefore not affecting the SROCC, we hope that this improves the readability and applicability of the HaarPSI. To also provide an objective measure of linear correlation, we repeated the numerical evaluation from Section~\ref{sec:results} with the Pearson product-moment correlation instead of the SROCC (see Table~\ref{tab:detailpearson} in Appendix~\ref{app:pearsoncorrelations}). The results of this analysis indicate that even without additional nonlinear regression, the HaarPSI has a highly linear relationship with human mean opinion scores from different databases and across varying types of distortion.

%haarpsi and fsim
The HaarPSI can conceptually be understood as a simplified version of the FSIM. Both metrics rely on the construction of two maps, where one map measures local similarities between a reference image and a distorted image and the other map assesses the relative importance of image areas. However, in the HaarPSI, these maps are defined only in terms of a single Haar wavelet filterbank, while the FSIM utilizes an implementation of the phase congruency measure that requires the input images to be convolved with 16 complex-valued filters and contains several non-trivial computational steps, like adaptive thresholding. Another difference is that the FSIM uses the phase congruency measure both as a weight function in {\eqref{eq:fsim}} and as a part of the local similarity measure {\eqref{eq:localfs}}. In the HaarPSI, the weight function {\eqref{eq:haarweight}} and the local similarity measure {\eqref{eq:localhs}} are strictly separated in the sense that they are based on distinct bands of the frequency spectrum.

These conceptual simplifications lead to a significant decrease in execution time (see Table~{\ref{tab:overallperf}}) and enable a better understanding of how single elements of the measure and properties of the input images contribute to the final similarity score. In the case of the HaarPSI, it is clear that the local similarity measure is based on high-frequency information, while the weight map, which provides a crude measure of visual saliency, is using filters that are tuned to lower frequencies. We suspect that a similar principle plays an important role in the FSIM, where additional high-frequency filters are applied to obtain the gradient map used in the local similarity measure {\eqref{eq:localfs}}. However, for the FSIM, it is difficult to verify this, as filters that are tuned to lower frequencies are only implicitly used in the computation of the phase congruency measure, which is in turn part of both the local similarity measure and the weight map.

We do not have a straightforward explanation as to why the HaarPSI outperforms the FSIM with respect to correlations with human opinion scores (see Table~{\ref{tab:sroccdatabases}}). After all, both measures have a similar overall structure and implement similar principles such as frequency and orientation selectivity. We assume that the reduced complexity of the HaarPSI also limits uncontrollable side effects when accentuating different aspects of the input images by varying the parameters $C$ and $\alpha$. This could improve the chance of successfully fitting subsets of benchmark databases when only considering two free parameters, but also decrease the generalization error. Furthermore, the principle of orientation selectivity is implemented differently in the HaarPSI in the sense that measurements regarding horizontal and vertical structures are only combined at the very end, that is, when taking the weighted average. It is well known that orientation selectivity is a strong organization principle in the primary visual cortex, where neurons that are tuned to similar orientations are grouped together in so-called orientation columns {\cite{hubel1974sequence}}. It thus seems reasonable that a consistent separation of the information yielded by vertical and horizontal filters has a positive effect on the correlations with human opinion scores. 

%other wavelets
From a computational point of view, it is very beneficial to apply discrete Haar wavelet filters instead of other wavelet filters. However, by changing $h^{\text{1D}}_1$ and $g^{\text{1D}}_1$ in \eqref{eq:haarfilters} to the respective filters, the measure given in \eqref{eq:haarpsi} can easily be defined for other wavelets. Table~\ref{tab:otherwavelets} depicts the performance of such measures based on selected Daubechies wavelets \cite{Daub1988}, symlets \cite{Daub1992}, coiflets \cite{Daub1993} and the Cohen-Daubechies-Feauveau wavelet \cite{CDF1992} with respect to the four databases considered in Section~\ref{sec:results}. It is interesting to see that Haar filters not only seem to be the computationally most efficient but also the qualitatively best choice for the measure \eqref{eq:haarpsi}.
% It is, however, possible that for specific distortions and with different
% choices for the constants $C_1$, $C_2$, and $\alpha$, changing the wavelet filter can increase the correlation with human opinion scores.
%It is, however, possible to fine-tune the presented method by distortion
%specific shoices for constants $C_1$, $C_2$, and $\alpha$ or by changing the wavelet
%filter in order to improve the performance for assess the quality of image
%degraded by a particular distortion type. 

\setlength{\tabcolsep}{2mm}
\begin{table}[!htb]
\centering
\caption{SROCC With Human Mean Opinion Scores for Different Wavelet Filters}
  \label{tab:otherwavelets}
\begin{threeparttable}
\begin{small}
\begin{tabular}{*{7}{c}}
\toprule[0.5mm]
\multicolumn{7}{c}{Grayscale Images}\\[0.1cm]
  & Daub2PSI & Daub4PSI & Sym4PSI & CDFPSI & Coif1PSI & HaarPSI\\
 LIVE & \cellcolor{green!25}0.9620 & \cellcolor{green!25}0.9530 & \cellcolor{green!25}0.9552 & \cellcolor{green!25}0.9604 & \cellcolor{green!25}0.9603 & \textbf{0.9690}\\
 TID2008 & 0.8971 & \cellcolor{green!25}0.8796 & 0.8915 & \cellcolor{green!25}0.8836 & 0.8965 & \textbf{0.9043}\\
 TID2013 & 0.8064 & 0.7982 & 0.8022 & 0.7965 & 0.8055 & \textbf{0.8094}\\
 CSIQ & 0.9492 & \cellcolor{green!25}0.9442 & 0.9454 & \cellcolor{green!25}0.9404 & 0.9485 & \textbf{0.9546}\\
\\
\multicolumn{7}{c}{Color Images}\\[0.1cm]
  & Daub2PSI & Daub4PSI & Sym4PSI & CDFPSI & Coif1PSI & HaarPSI\\
 LIVE & 0.9659 & \cellcolor{green!25}0.9610 & 0.9630 & 0.9675 & 0.9644 & \textbf{0.9683}\\
 TID2008 & 0.8992 & \cellcolor{green!25}0.8804 & \cellcolor{green!25}0.8950 & \cellcolor{green!25}0.8932 & 0.8986 & \textbf{0.9097}\\
 TID2013 & 0.8724 & 0.8643 & 0.8696 & 0.8633 & 0.8716 & \textbf{0.8732}\\
 CSIQ & 0.9603 & 0.9577 & 0.9592 & 0.9596 & 0.9593 & \textbf{0.9604}\\\midrule[0.5mm]
\end{tabular}
\end{small}
\begin{scriptsize}
\begin{tablenotes}
\item \colorbox{green!25}{Lower correlation than HaarPSI. The difference is statistically significant with $p < 0.05$.}
\item \colorbox{red!25}{Higher correlation than HaarPSI. The difference is statistically significant with $p < 0.05$.}
\item The highest correlation in each row is written in \textbf{boldface}.
\end{tablenotes}
\end{scriptsize}
\end{threeparttable}

\end{table}

A Matlab function implementing the HaarPSI can be downloaded from \url{www.haarpsi.org}.
\FloatBarrier
\section*{Acknowledgements}
{R. Reisenhofer would like to thank Eero Simoncelli, Johannes Ball\'{e} and the members of the Laboratory for Computational Vision at NYU for their kind hospitality and very insightful discussions. S. Bosse and G. Kutyniok would like to thank Anthony Norcia for fruitful discussions. G. Kutyniok would also like to thank Eero Simoncelli for interesting discussions on the topic and to acknowledge support by the Einstein Foundation Berlin, the Einstein Center for Mathematics Berlin (ECMath), the European Commission-Project DEDALE (contract no. 665044) within the H2020 Framework Program, DFG Grant KU 1446/18, DFG-SPP 1798 Grants KU 1446/21 and KU 1446/23, the DFG Collaborative Research Center TRR 109 Discretization in Geometry and Dynamics, and by the DFG Research Center Matheon Mathematics for Key Technologies in Berlin.}

\bibliographystyle{unsrt}
\bibliography{refsHaarPSI}

\begin{thebibliography}{10}

\bibitem{Cisco}
Cisco.
\newblock Cisco visual networking index: Forecast and methodology,
  {2015–-2020}.
\newblock {\em White paper}, 2016.

\bibitem{Lin2011}
W.~Lin and C.-C.~J. Kuo.
\newblock {Perceptual visual quality metrics: A survey}.
\newblock {\em Journal of Visual Communication and Image Representation},
  22(4):297--312, 2011.

\bibitem{Gir93}
B.~Girod.
\newblock {What's wrong with mean-squared error?}
\newblock In {\em Digital Images and Human Vision}, pages 207--220. 1993.

\bibitem{Watsona1997}
A.B. Watson, R.~Borthwick, and M.~Taylor.
\newblock {Image quality and entropy masking}.
\newblock In {\em SPIE Proceedings}, volume 3016, pages 1--11, 1997.

\bibitem{daly1990}
Scott~J Daly.
\newblock {Application of a noise-adaptive contrast sensitivity function to
  image data compression}.
\newblock {\em Optical Engineering}, 29(8):977--987, 1990.

\bibitem{Lubin1997}
Jeffrey Lubin.
\newblock {A human vision system model for objective picture quality
  measurements}.
\newblock {\em International Broadcasting Convention}, pages 498--503, 1997.

\bibitem{jia2006estimating}
Yuting Jia, Weisi Lin, and Ashraf~A Kassim.
\newblock Estimating just-noticeable distortion for video.
\newblock {\em Circuits and Systems for Video Technology, IEEE Transactions
  on}, 16(7):820--829, 2006.

\bibitem{LaCh2010}
Eric~Cooper Larson and Damon~Michael Chandler.
\newblock Most apparent distortion: Full-reference image quality assessment and
  the role of strategy.
\newblock {\em Journal of Electronic Imaging}, 19(1):011006--1--011006--21,
  2010.

\bibitem{WBSS2004}
Zhou Wang, Alan~C. Bovik, Hamid~R. Sheikh, and Eero~P. Simoncelli.
\newblock Image quality assessment: From error visibility to structural
  similarity.
\newblock {\em IEEE Trans. Image Proc.}, 13(4):600--612, 2004.

\bibitem{ShBo2006}
H.~R. Sheikh and A.~C. Bovik.
\newblock Image information and visual quality.
\newblock {\em IEEE Transactions on Image Processing}, 15:430--444, 2006.

\bibitem{LLN2012}
A.~Liu, W.~Lin, and M.~Narwaria.
\newblock Image quality assessment based on gradient similarity.
\newblock {\em IEEE Transactions on Image Processing}, 21(4):1500--1512, April
  2012.

\bibitem{ZhLi2012}
L.~Zhang and H.~Li.
\newblock {SR-SIM}: A fast and high performance {IQA} index based on spectral
  residual.
\newblock In {\em 2012 19th IEEE International Conference on Image Processing},
  pages 1473--1476, Sept 2012.

\bibitem{ZSL2014}
L.~Zhang, Y.~Shen, and H.~Li.
\newblock Vsi: A visual saliency-induced index for perceptual image quality
  assessment.
\newblock {\em IEEE Transactions on Image Processing}, 23(10):4270--4281, Oct
  2014.

\bibitem{DoYa2011}
Richard Dosselmann and Xue~Dong Yang.
\newblock A comprehensive assessment of the structural similarity index.
\newblock {\em Signal, Image and Video Processing}, 5(1):81--91, 2011.

\bibitem{WSB2003}
Zhou Wang, Eero~P. Simoncelli, and Alan~C. Bovik.
\newblock Multi-scale structural similarity fror image quality assessment.
\newblock In {\em Proceedings of 37th IEEE Asilomar Conference on Signals,
  Systems and Computers}, 2003.

\bibitem{ZZMZ2011}
Lin Zhang, Lei Zhang, Xuanqin Mou, and David Zhang.
\newblock {FSIM}: A feature similarity index for image quality assessment.
\newblock {\em IEEE Trans. Image Proc.}, 20(8):2378--2386, 2011.

\bibitem{Kang2014}
L.~Kang, P.~Ye, Y.~Li, and D.~Doermann.
\newblock {Convolutional neural networks for no-reference image quality
  assessment}.
\newblock In {\em Computer Vision and Pattern Recognition (CVPR), 2014 IEEE
  Conference on}, pages 1733--1740, 2014.

\bibitem{Ye2012}
P.~Ye and D.~Doermann.
\newblock {No-reference image quality assessment using visual codebooks}.
\newblock {\em IEEE Transactions on Image Processing}, 21(7):3129--3138, 2012.

\bibitem{Zhang2015}
P.~Zhang, W.~Zhou, L.~Wu, and H.~Li.
\newblock {SOM: Semantic obviousness metric for image quality assessment}.
\newblock {\em 2015 IEEE Conference on Computer Vision and Pattern Recognition
  (CVPR)}, pages 2394--2402, 2015.

\bibitem{bosse2016dnnNrIqa}
S.~Bosse, D.~Maniry, T.~Wiegand, and W.~Samek.
\newblock A deep neural network for image quality assessment.
\newblock In {\em Image Processing (ICIP), 2016 IEEE International Conference
  on}, 2016.

\bibitem{bosse2016dnnFrIqa}
S.~Bosse, D.~Maniry, K.-R. M\"uller, T.~Wiegand, and W.~Samek.
\newblock Full-reference image quality assessment using neural networks.
\newblock In {\em Int. Work. Qual. Multimed. Exp.}, 2016.

\bibitem{Kov2000}
Peter Kovesi.
\newblock Phase congruency: A low-level image invariant.
\newblock {\em Psychological Research}, 64:136--148, 2000.

\bibitem{KovONLINE}
Peter~D. Kovesi.
\newblock Matlab and octave functions for computer vision and image processing.
\newblock Centre for Exploration Targeting, School of Earth and Environment,
  The University of Western Australia.
\newblock Available from
  \url{http://www.csse.uwa.edu.au/~pk/research/matlabfns/}.

\bibitem{MRBO1986}
M.~C. Morrone, J.~R. Ross, D.~C. Burr, and R.~A. Owens.
\newblock Mach bands are phase dependent.
\newblock {\em Nature}, 324(6094):250--253, 1986.

\bibitem{Haar1910}
Alfred Haar.
\newblock Zur {T}heorie der orthogonalen {F}unktionensysteme.
\newblock {\em Mathematische Annalen}, 69(3):331--371, 1910.

\bibitem{PLZECB2009}
N.~Ponomarenko, V.~Lukin, A.~Zelensky, K.~Egiazarian, M.~Carli, and
  F.~Battisti.
\newblock {TID2008} - a database for evaluation of full-reference visual
  quality assessment metrics.
\newblock {\em Advances of Modern Radioelectronics}, 10:30--45, 2009.

\bibitem{Ponomarenko2015}
Nikolay Ponomarenko, Lina Jin, Oleg Ieremeiev, Vladimir Lukin, Karen
  Egiazarian, Jaakko Astola, Benoit Vozel, Kacem Chehdi, Marco Carli, Federica
  Battisti, and C.-C.~Jay Kuo.
\newblock Image database {TID2013}: Peculiarities, results and perspectives.
\newblock {\em Signal Processing: Image Communication}, 30:57 -- 77, 2015.

\bibitem{SWCBOnline}
Hamid~Rahim Sheikh, Zhou Wang, Lawrance Cormack, and Alan~C. Bovik.
\newblock {LIVE} image quality assessment database release 2.
\newblock Available from \url{http://live.ece.utexas.edu/research/quality}.

\bibitem{ITUTP1401}
International~Telecommunication Union.
\newblock {ITU-T P.1401}, methods, metrics and procedures for statistical
  evaluation, qualification and comparison of objective quality prediction
  models.
\newblock 2012.

\bibitem{fieller1957tests}
Edgar~C Fieller, Herman~O Hartley, and Egon~S Pearson.
\newblock Tests for rank correlation coefficients. {I}.
\newblock {\em Biometrika}, 44(3/4):470--481, 1957.

\bibitem{palmer1999vision}
S.~E. Palmer.
\newblock {\em {Vision science: Photons to phenomenology}}, volume~1.
\newblock MIT press Cambridge, MA, 1999.

\bibitem{laparra2017perceptually}
Valero Laparra, Alex Berardino, Johannes Ball{\'e}, and Eero~P Simoncelli.
\newblock Perceptually optimized image rendering.
\newblock {\em arXiv preprint arXiv:1701.06641}, 2017.

\bibitem{hubel1974sequence}
David~H Hubel and Torsten~N Wiesel.
\newblock Sequence regularity and geometry of orientation columns in the monkey
  striate cortex.
\newblock {\em Journal of Comparative Neurology}, 158(3):267--293, 1974.

\bibitem{Daub1988}
Ingrid Daubechies.
\newblock Orthonormal bases of compactly supported wavelets.
\newblock {\em Communications on Pure and Applied Mathematics}, 41(7):909--996,
  1988.

\bibitem{Daub1992}
Ingrid Daubechies.
\newblock {\em Ten Lectures on Wavelets}.
\newblock Society for Industrial and Applied Mathematics, 1992.

\bibitem{Daub1993}
Ingrid Daubechies.
\newblock Orthonormal bases of compactly supported wavelets {II}: Variations on
  a theme.
\newblock {\em SIAM J. Math. Anal.}, 24(2):499--519, 1993.

\bibitem{CDF1992}
A.~Cohen, Ingrid Daubechies, and J.-C. Feauveau.
\newblock Biorthogonal bases of compactly supported wavelets.
\newblock {\em Communications on Pure and Applied Mathematics}, 45(5):485--560,
  1992.

\end{thebibliography}

\clearpage

\appendix
\section{Pearson Product-Moment Correlations}
\label{app:pearsoncorrelations}
\begin{table}[!htb]
\centering
\caption{Pearson Correlations of IQA Metrics With Human Mean Opinion Scores}
 \label{tab:detailpearson}
\begin{scriptsize}
	\resizebox{0.9\columnwidth}{!}{
\begin{threeparttable}
	
\begin{tabular}{*{12}{c}}
\toprule[0.5mm]
\multicolumn{12}{c}{Color Images}\\[0.1cm]
 & & PSNR & VIF & SSIM & MSSSIM & GSM & MAD & SRSIM & FSIM & VSI & HaarPSI\\
& LIVE & \cellcolor{green!25}0.8585 & \cellcolor{green!25}0.9411 & \cellcolor{green!25}0.8290 & \cellcolor{green!25}0.7670 & \cellcolor{green!25}0.7799 & 0.9559 & \cellcolor{green!25}0.7758 & \cellcolor{green!25}0.8595 & \cellcolor{green!25}0.7647 & \textbf{0.9592}\\
& TID2008 & \cellcolor{green!25}0.5190 & \cellcolor{green!25}0.7769 & \cellcolor{green!25}0.7401 & \cellcolor{green!25}0.7897 & \cellcolor{green!25}0.7779 & \cellcolor{green!25}0.8290 & \cellcolor{green!25}0.8242 & \cellcolor{green!25}0.8341 & \cellcolor{green!25}0.8107 & \textbf{0.9032}\\
& TID2013 & \cellcolor{green!25}0.4785 & \cellcolor{green!25}0.7335 & \cellcolor{green!25}0.7596 & \cellcolor{green!25}0.7773 & \cellcolor{green!25}0.7966 & \cellcolor{green!25}0.8074 & \cellcolor{green!25}0.7984 & \cellcolor{green!25}0.8322 & \cellcolor{green!25}0.8373 & \textbf{0.8904}\\
& CSIQ & \cellcolor{green!25}0.7512 & \cellcolor{green!25}0.9219 & \cellcolor{green!25}0.7916 & \cellcolor{green!25}0.7720 & \cellcolor{green!25}0.7471 & \textbf{0.9500} & \cellcolor{green!25}0.7520 & \cellcolor{green!25}0.8208 & \cellcolor{green!25}0.8392 & 0.9463\\
\\
\multicolumn{12}{c}{Color Images}\\[0.1cm]
\multirow{5}{*}{LIVE} & jpg2k & \cellcolor{green!25}0.8747 & \cellcolor{green!25}0.9476 & \cellcolor{green!25}0.8925 & \cellcolor{green!25}0.8697 & \cellcolor{green!25}0.8564 & \textbf{0.9725} & \cellcolor{green!25}0.8800 & \cellcolor{green!25}0.9036 & \cellcolor{green!25}0.8662 & 0.9673\\
& jpg & \cellcolor{green!25}0.8650 & \cellcolor{green!25}0.9600 & \cellcolor{green!25}0.9279 & \cellcolor{green!25}0.9184 & \cellcolor{green!25}0.9131 & 0.9742 & \cellcolor{green!25}0.9028 & \cellcolor{green!25}0.9117 & \cellcolor{green!25}0.9037 & \textbf{0.9779}\\
& gwn & \textbf{0.9792} & \cellcolor{green!25}0.9632 & \cellcolor{green!25}0.9583 & \cellcolor{green!25}0.9181 & \cellcolor{green!25}0.8904 & 0.9764 & \cellcolor{green!25}0.8684 & \cellcolor{green!25}0.9263 & \cellcolor{green!25}0.9171 & 0.9791\\
& gblur & \cellcolor{green!25}0.7744 & 0.9575 & \cellcolor{green!25}0.8881 & \cellcolor{green!25}0.8450 & \cellcolor{green!25}0.8565 & 0.9486 & \cellcolor{green!25}0.8411 & \cellcolor{green!25}0.9086 & \cellcolor{green!25}0.8544 & \textbf{0.9576}\\
& ff & \cellcolor{green!25}0.8753 & \textbf{0.9560} & \cellcolor{green!25}0.8619 & \cellcolor{green!25}0.8113 & \cellcolor{green!25}0.7925 & 0.9461 & \cellcolor{green!25}0.7837 & \cellcolor{green!25}0.8515 & \cellcolor{green!25}0.8151 & 0.9444\\
\\
\multirow{17}{*}{TID2008} & gwn & \textbf{0.9336} & 0.8657 & \cellcolor{green!25}0.7494 & \cellcolor{green!25}0.7433 & \cellcolor{green!25}0.8078 & \cellcolor{green!25}0.8165 & \cellcolor{green!25}0.8284 & \cellcolor{green!25}0.8076 & 0.8719 & 0.9029\\
& gwnc & \textbf{0.9208} & 0.8928 & \cellcolor{green!25}0.7758 & \cellcolor{green!25}0.7772 & \cellcolor{green!25}0.7833 & \cellcolor{green!25}0.8267 & 0.8625 & 0.8671 & 0.9045 & 0.9131\\
& scn & \textbf{0.9526} & \cellcolor{green!25}0.8578 & \cellcolor{green!25}0.7678 & \cellcolor{green!25}0.7583 & \cellcolor{green!25}0.8422 & \cellcolor{green!25}0.8598 & \cellcolor{green!25}0.8492 & \cellcolor{green!25}0.8217 & 0.8862 & 0.9283\\
& mn & \cellcolor{red!25}0.8627 & \cellcolor{red!25}\textbf{0.8900} & 0.7496 & 0.7849 & \cellcolor{green!25}0.5512 & 0.7566 & 0.7345 & 0.8106 & 0.6114 & 0.7480\\
& hfn & \cellcolor{red!25}\textbf{0.9680} & 0.9441 & \cellcolor{green!25}0.8228 & \cellcolor{green!25}0.8176 & \cellcolor{green!25}0.8452 & \cellcolor{green!25}0.8931 & \cellcolor{green!25}0.8657 & \cellcolor{green!25}0.8597 & \cellcolor{green!25}0.8934 & 0.9393\\
& in & \textbf{0.8566} & 0.8146 & \cellcolor{green!25}0.6202 & \cellcolor{green!25}0.6220 & \cellcolor{green!25}0.6218 & \cellcolor{green!25}0.0417 & 0.6912 & 0.7044 & 0.7651 & 0.8077\\
& qn & \textbf{0.8729} & \cellcolor{green!25}0.7442 & \cellcolor{green!25}0.7239 & \cellcolor{green!25}0.7602 & 0.8090 & 0.7981 & \cellcolor{green!25}0.7586 & 0.7986 & 0.8077 & 0.8602\\
& gblr & 0.8439 & \cellcolor{red!25}\textbf{0.9388} & 0.8936 & 0.8745 & 0.8761 & 0.9227 & 0.9078 & 0.9078 & 0.8731 & 0.8934\\
& den & \cellcolor{green!25}0.9428 & \cellcolor{green!25}0.8968 & \cellcolor{green!25}0.9208 & \cellcolor{green!25}0.9156 & \cellcolor{green!25}0.9052 & 0.9612 & \cellcolor{green!25}0.9133 & \cellcolor{green!25}0.9344 & \cellcolor{green!25}0.9162 & \textbf{0.9739}\\
& jpg & \cellcolor{green!25}0.8597 & \cellcolor{green!25}0.9327 & \cellcolor{green!25}0.9319 & \cellcolor{green!25}0.9279 & 0.9546 & 0.9487 & 0.9444 & \cellcolor{green!25}0.9299 & 0.9566 & \textbf{0.9647}\\
& jpg2k & \cellcolor{green!25}0.8629 & \cellcolor{green!25}0.9169 & \cellcolor{green!25}0.9492 & \cellcolor{green!25}0.9365 & \cellcolor{green!25}0.9564 & \cellcolor{green!25}0.9733 & \cellcolor{green!25}0.8965 & \cellcolor{green!25}0.9566 & \cellcolor{green!25}0.9632 & \textbf{0.9856}\\
& jpgt & \cellcolor{green!25}0.6258 & 0.8720 & 0.8375 & 0.8150 & 0.8441 & 0.8556 & 0.8573 & 0.8446 & 0.8705 & \textbf{0.8882}\\
& jpg2kt & 0.8528 & 0.8307 & 0.8252 & 0.7970 & 0.7958 & 0.8295 & 0.7932 & 0.7883 & 0.8142 & \textbf{0.8688}\\
& pn & \cellcolor{green!25}0.5831 & 0.7366 & 0.6685 & \cellcolor{green!25}0.6637 & 0.7013 & \textbf{0.8242} & 0.7381 & 0.7297 & 0.7314 & 0.7936\\
& bdist & \cellcolor{green!25}0.6277 & 0.8340 & 0.8659 & 0.7861 & \textbf{0.8822} & 0.8007 & 0.7864 & 0.8410 & \cellcolor{green!25}0.6198 & 0.8069\\
& ms & 0.6845 & 0.5896 & 0.6834 & 0.6735 & \cellcolor{red!25}\textbf{0.7431} & 0.5709 & 0.6098 & 0.6700 & 0.6420 & 0.5358\\
& ctrst & 0.5819 & \cellcolor{red!25}\textbf{0.8816} & 0.5158 & 0.7686 & 0.7068 & \cellcolor{green!25}0.2573 & 0.6978 & 0.7275 & 0.6995 & 0.6446\\
\\
\multirow{24}{*}{TID2013} & gwn & \textbf{0.9519} & 0.9010 & \cellcolor{green!25}0.7954 & \cellcolor{green!25}0.7891 & \cellcolor{green!25}0.8500 & \cellcolor{green!25}0.8732 & \cellcolor{green!25}0.8569 & \cellcolor{green!25}0.8435 & 0.8928 & 0.9248\\
& gwnc & 0.8948 & 0.8641 & \cellcolor{green!25}0.7615 & \cellcolor{green!25}0.7629 & \cellcolor{green!25}0.8216 & \cellcolor{green!25}0.8297 & 0.8603 & 0.8543 & 0.8975 & \textbf{0.8998}\\
& scn & \textbf{0.9513} & \cellcolor{green!25}0.8783 & \cellcolor{green!25}0.7840 & \cellcolor{green!25}0.7681 & \cellcolor{green!25}0.8420 & \cellcolor{green!25}0.8804 & \cellcolor{green!25}0.8371 & \cellcolor{green!25}0.8240 & \cellcolor{green!25}0.8714 & 0.9261\\
& mn & 0.8447 & \cellcolor{red!25}\textbf{0.8772} & 0.7569 & 0.7929 & \cellcolor{green!25}0.5934 & 0.7804 & 0.7615 & 0.8214 & 0.6585 & 0.7737\\
& hfn & \textbf{0.9607} & 0.9454 & \cellcolor{green!25}0.8342 & \cellcolor{green!25}0.8307 & \cellcolor{green!25}0.8575 & 0.9098 & \cellcolor{green!25}0.8702 & \cellcolor{green!25}0.8669 & \cellcolor{green!25}0.8939 & 0.9415\\
& in & \textbf{0.8856} & 0.8489 & \cellcolor{green!25}0.6625 & \cellcolor{green!25}0.6541 & \cellcolor{green!25}0.6602 & \cellcolor{green!25}0.2741 & \cellcolor{green!25}0.7183 & \cellcolor{green!25}0.7216 & 0.7776 & 0.8325\\
& qn & \textbf{0.8855} & \cellcolor{green!25}0.7805 & \cellcolor{green!25}0.7514 & \cellcolor{green!25}0.7752 & 0.8199 & 0.8365 & \cellcolor{green!25}0.7677 & 0.8096 & 0.8119 & 0.8643\\
& gblr & 0.8952 & \cellcolor{red!25}\textbf{0.9530} & 0.8832 & 0.8616 & 0.8565 & 0.9336 & 0.8893 & 0.8922 & 0.8548 & 0.9030\\
& den & 0.9572 & \cellcolor{green!25}0.8914 & \cellcolor{green!25}0.9199 & \cellcolor{green!25}0.9110 & \cellcolor{green!25}0.9116 & 0.9602 & \cellcolor{green!25}0.9114 & \cellcolor{green!25}0.9304 & \cellcolor{green!25}0.9187 & \textbf{0.9690}\\
& jpg & \cellcolor{green!25}0.8972 & \cellcolor{green!25}0.9332 & \cellcolor{green!25}0.9278 & \cellcolor{green!25}0.9207 & \cellcolor{green!25}0.9470 & \cellcolor{green!25}0.9510 & \cellcolor{green!25}0.9343 & \cellcolor{green!25}0.9242 & \cellcolor{green!25}0.9479 & \textbf{0.9750}\\
& jpg2k & \cellcolor{green!25}0.9078 & \cellcolor{green!25}0.9184 & \cellcolor{green!25}0.9424 & \cellcolor{green!25}0.9183 & \cellcolor{green!25}0.9462 & 0.9663 & \cellcolor{green!25}0.8772 & \cellcolor{green!25}0.9360 & \cellcolor{green!25}0.9494 & \textbf{0.9787}\\
& jpgt & \cellcolor{green!25}0.6410 & 0.9000 & 0.8721 & \cellcolor{green!25}0.8476 & 0.8697 & \cellcolor{green!25}0.8537 & 0.8772 & 0.8761 & 0.8972 & \textbf{0.9177}\\
& jpg2kt & 0.8834 & 0.8692 & \cellcolor{green!25}0.8260 & \cellcolor{green!25}0.7929 & \cellcolor{green!25}0.7960 & 0.8648 & \cellcolor{green!25}0.7914 & \cellcolor{green!25}0.8010 & \cellcolor{green!25}0.8179 & \textbf{0.8913}\\
& pn & \cellcolor{green!25}0.6702 & 0.7686 & 0.7481 & \cellcolor{green!25}0.7376 & 0.7718 & \textbf{0.8513} & 0.8034 & 0.7957 & 0.7971 & 0.8376\\
& bdist & \cellcolor{green!25}0.1448 & 0.5027 & 0.5589 & 0.4608 & \textbf{0.5939} & 0.3184 & 0.4436 & 0.5237 & \cellcolor{green!25}0.1356 & 0.4441\\
& ms & 0.7482 & 0.6829 & 0.7309 & 0.6823 & \cellcolor{red!25}\textbf{0.8153} & 0.6654 & 0.6364 & 0.7103 & 0.7367 & 0.6365\\
& ctrst & 0.4812 & \cellcolor{red!25}\textbf{0.8730} & 0.4941 & 0.7268 & 0.6701 & \cellcolor{green!25}0.2601 & 0.6520 & 0.6838 & 0.6595 & 0.5916\\
& ccs & \cellcolor{green!25}0.1378 & \cellcolor{green!25}0.3404 & 0.4349 & 0.4237 & \cellcolor{green!25}0.3739 & \cellcolor{green!25}0.0351 & \cellcolor{green!25}0.2491 & 0.6069 & \textbf{0.6852} & 0.6003\\
& mgn & \textbf{0.9187} & 0.8559 & \cellcolor{green!25}0.7358 & \cellcolor{green!25}0.7301 & \cellcolor{green!25}0.7903 & 0.8422 & \cellcolor{green!25}0.8049 & \cellcolor{green!25}0.8008 & 0.8505 & 0.8786\\
& cn & \cellcolor{green!25}0.8548 & \cellcolor{green!25}0.8992 & \cellcolor{green!25}0.8459 & \cellcolor{green!25}0.8105 & \cellcolor{green!25}0.9286 & \cellcolor{green!25}0.9280 & \cellcolor{green!25}0.9260 & \cellcolor{green!25}0.9214 & \cellcolor{green!25}0.9301 & \textbf{0.9571}\\
& lcni & \cellcolor{green!25}0.9372 & \cellcolor{green!25}0.9034 & \cellcolor{green!25}0.9058 & \cellcolor{green!25}0.8917 & \cellcolor{green!25}0.9472 & 0.9520 & \cellcolor{green!25}0.9439 & \cellcolor{green!25}0.9364 & \cellcolor{green!25}0.9463 & \textbf{0.9686}\\
& icqd & \textbf{0.9227} & 0.8582 & \cellcolor{green!25}0.8083 & \cellcolor{green!25}0.7767 & 0.8240 & 0.8626 & \cellcolor{green!25}0.7574 & \cellcolor{green!25}0.8053 & \cellcolor{green!25}0.8083 & 0.8826\\
& cha & \cellcolor{green!25}0.8569 & 0.9441 & 0.9519 & \cellcolor{green!25}0.9071 & \textbf{0.9563} & 0.9560 & \cellcolor{green!25}0.8819 & 0.9478 & 0.9498 & 0.9549\\
& ssr & \cellcolor{green!25}0.9167 & \cellcolor{green!25}0.9067 & \cellcolor{green!25}0.9528 & \cellcolor{green!25}0.9197 & \cellcolor{green!25}0.9601 & 0.9658 & \cellcolor{green!25}0.9135 & \cellcolor{green!25}0.9412 & \cellcolor{green!25}0.9449 & \textbf{0.9791}\\
\\
\multirow{6}{*}{CSIQ} & gwn & 0.9437 & \textbf{0.9590} & \cellcolor{green!25}0.8043 & \cellcolor{green!25}0.8254 & \cellcolor{green!25}0.8517 & 0.9486 & \cellcolor{green!25}0.8669 & \cellcolor{green!25}0.7959 & \cellcolor{green!25}0.8875 & 0.9433\\
& jpeg & \cellcolor{green!25}0.7898 & \cellcolor{green!25}0.9590 & \cellcolor{green!25}0.9165 & \cellcolor{green!25}0.9064 & \cellcolor{green!25}0.8964 & 0.9696 & \cellcolor{green!25}0.8731 & \cellcolor{green!25}0.9077 & \cellcolor{green!25}0.8833 & \textbf{0.9780}\\
& jpg2k & \cellcolor{green!25}0.9270 & \cellcolor{green!25}0.9360 & \cellcolor{green!25}0.8967 & \cellcolor{green!25}0.8843 & \cellcolor{green!25}0.8793 & 0.9808 & \cellcolor{green!25}0.8428 & \cellcolor{green!25}0.9106 & \cellcolor{green!25}0.9008 & \textbf{0.9853}\\
& gpn & 0.9527 & \textbf{0.9552} & \cellcolor{green!25}0.7844 & \cellcolor{green!25}0.7790 & \cellcolor{green!25}0.8293 & 0.9548 & \cellcolor{green!25}0.7777 & \cellcolor{green!25}0.8160 & \cellcolor{green!25}0.8698 & 0.9470\\
& gblr & \cellcolor{green!25}0.9081 & 0.9627 & \cellcolor{green!25}0.8692 & \cellcolor{green!25}0.8670 & \cellcolor{green!25}0.8575 & \textbf{0.9713} & \cellcolor{green!25}0.8675 & \cellcolor{green!25}0.8843 & \cellcolor{green!25}0.8761 & 0.9623\\
& ctrst & 0.8888 & 0.9294 & \cellcolor{green!25}0.7666 & 0.9003 & \cellcolor{green!25}0.8656 & \textbf{0.9306} & 0.8878 & 0.8765 & \cellcolor{green!25}0.8686 & 0.9229\\\midrule[0.5mm]
\end{tabular}
\begin{tablenotes}
\item \colorbox{green!25}{Lower correlation than HaarPSI. The difference is statistically significant with $p < 0.05$.}
\item \colorbox{red!25}{Higher correlation than HaarPSI. The difference is statistically significant with $p < 0.05$.}
\item The highest correlation in each row is written in \textbf{boldface}.
\item All correlations were obtained \textbf{without nonlinear regression}.
\end{tablenotes}
\end{threeparttable}}
\end{scriptsize}
\end{table}

\end{document}